\documentclass{article} 
\usepackage{iclr2025_conference,times}

\usepackage{amsmath,amsfonts,bm}

\def\eqref#1{equation~\ref{#1}}

\def\1{\bm{1}}

\DeclareMathAlphabet{\mathsfit}{\encodingdefault}{\sfdefault}{m}{sl}
\SetMathAlphabet{\mathsfit}{bold}{\encodingdefault}{\sfdefault}{bx}{n}

\usepackage{xparse}
\NewDocumentCommand{\bang}{ mO{} }{\textcolor{blue}{\textsuperscript{\textit{bang}}\textsf{\textbf{\small[#1]}}}}

\NewDocumentCommand{\Charles}{ mO{} }{\textcolor{blue}{\textsuperscript{\textit{Charles}}\textsf{\textbf{\small[#1]}}}}

\usepackage{hyperref}
\usepackage{url}
\usepackage[utf8]{inputenc} 
\usepackage[T1]{fontenc}    
\usepackage{hyperref}       
\usepackage{url}            
\usepackage{booktabs}       
\usepackage{amsfonts}       
\usepackage{nicefrac}       
\usepackage{microtype}      
\usepackage{xcolor}         

\usepackage{times}  
\usepackage{helvet}  
\usepackage{courier}  
\usepackage{graphicx} 
\usepackage{booktabs}
\usepackage{algorithm}
\usepackage{algpseudocode}
\usepackage{multirow, makecell}
\usepackage{soul}
\usepackage[normalem]{ulem}
\usepackage{xcolor}
\usepackage[most,breakable]{tcolorbox}
\usepackage{xcolor}
\usepackage{colortbl}

\usepackage{booktabs}
\usepackage{caption}

\usepackage{array} 

\usepackage{CJK}
\usepackage{longtable} 
\usepackage{lscape}

\usepackage{courier}  

\usepackage{bibentry}
\usepackage{cleveref}
\usepackage{listings}

\usepackage{algorithm}  
\usepackage{algpseudocode}  
\usepackage{amsmath} 

\usepackage{fancyhdr}
\renewcommand{\headrulewidth}{0pt}  
\newcommand{\tablestyle}[2]{\setlength{\tabcolsep}{#1}\renewcommand{\arraystretch}{#2}\centering\footnotesize}

\newcommand{\model}{Data Interpreter}

\definecolor{royalblue(web)}{rgb}{0.25, 0.41, 0.88}
\definecolor{blue-violet}{rgb}{0.54, 0.17, 0.89}
\definecolor{brightmaroon}{rgb}{0.76, 0.13, 0.28}
\definecolor{darkmagenta}{rgb}{0.55, 0.0, 0.55}
\definecolor{bleudefrance}{rgb}{0.19, 0.55, 0.91}
\definecolor{palatinateblue}{rgb}{0.15, 0.23, 0.89}
\definecolor{royalblue(web)}{rgb}{0.25, 0.41, 0.88}
\definecolor{whitesmoke}{rgb}{0.96, 0.96, 0.96}
\definecolor{thulianpink}{rgb}{0.87, 0.44, 0.63}
\definecolor{amber(sae/ece)}{rgb}{1.0, 0.49, 0.0}
\definecolor{darkblue}{rgb}{0.0, 0.0, 0.55}
\definecolor{alizarin}{rgb}{0.82, 0.1, 0.26}
\definecolor{asparagus}{rgb}{0.53, 0.66, 0.42}
\definecolor{darkspringgreen}{rgb}{0.09, 0.45, 0.27}
\definecolor{columbiablue}{rgb}{0.61, 0.87, 1.0}
\definecolor{wildblueyonder}{rgb}{0.64, 0.68, 0.82}
\definecolor{trolleygrey}{rgb}{0.5, 0.5, 0.5}
\definecolor{paleaqua}{rgb}{0.74, 0.83, 0.9}
\definecolor{bubblegum}{rgb}{0.99, 0.76, 0.8}
\definecolor{coralred}{rgb}{1.0, 0.25, 0.25}
\definecolor{green(ryb)}{rgb}{0.4, 0.69, 0.2}
\definecolor{flame}{rgb}{0.89, 0.35, 0.13}
\definecolor{bittersweet}{rgb}{1.0, 0.44, 0.37}
\definecolor{darksalmon}{rgb}{0.91, 0.59, 0.48}
\definecolor{emerald}{rgb}{0.31, 0.78, 0.47}
\definecolor{green(pigment)}{rgb}{0.0, 0.65, 0.31}

\definecolor{codegreen}{rgb}{0,0.6,0}
\definecolor{codegray}{rgb}{0.5,0.5,0.5}
\definecolor{codepurple}{rgb}{0.58,0,0.82}
\definecolor{backcolour}{rgb}{0.96,0.96,0.94}
\definecolor{bluegray}{rgb}{0.3, 0.38, 0.47}
\definecolor{whitesmoke}{rgb}{0.96, 0.96, 0.96}
\definecolor{codegreen}{rgb}{0,0.6,0}
\definecolor{codegray}{rgb}{0.5,0.5,0.5}
\definecolor{codepurple}{rgb}{0.58,0,0.82}
\definecolor{backcolour}{rgb}{0.96,0.96,0.94}

\NewDocumentCommand{\mc}{ mO{} }
{
  \textcolor{red}{\textsuperscript{Mingchen}}
  \textcolor{blue}{\textsf{\textbf{[#1]}}}
}

\title{Data Interpreter: \\An LLM Agent For
Data Science}

\author{%
 \textbf{Sirui Hong}$^1 \footnotemark[1]$ ,
 \textbf{Yizhang Lin}$^1 \thanks{These authors contributed equally to this work.}$,  
 \textbf{Bang Liu}$^{2\S}\footnotemark[2]$, 
 \textbf{Bangbang Liu}$^1 \thanks{The authors are listed in alphabetical order.}$,
 \textbf{Binhao Wu}$^1 \footnotemark[2]$,
 \textbf{Ceyao Zhang}$^{3} \footnotemark[2]$, \\
 \textbf{Chenxing Wei}$^{4} \footnotemark[2],$
 \textbf{Danyang Li}$^1 \footnotemark[2]$, 
 \textbf{Jiaqi Chen}$^5 \footnotemark[2]$, 
 \textbf{Jiayi Zhang}$^6 \footnotemark[2]$,
 \textbf{Jinlin Wang}$^1 \footnotemark[2]$,
 \textbf{Li Zhang}$^5 \footnotemark[2]$,\\
  \textbf{Lingyao Zhang} \footnotemark[2],
  \textbf{Min Yang}$^7\footnotemark[2]$,
  \textbf{Mingchen Zhuge}$^8 \footnotemark[2]$,
  \textbf{Taicheng Guo}$^9 \footnotemark[2]$,
  \textbf{Tuo Zhou}$^{10} \footnotemark[2]$,
  \textbf{Wei Tao}$^5 \footnotemark[2]$,\\
  \textbf{Xiangru Tang}$^{11} \footnotemark[2]$,
  \textbf{Xiangtao Lu}$^1 \footnotemark[2]$,
  \textbf{Xiawu Zheng}$^{12} \footnotemark[2]$,
  \textbf{Xinbing Liang}$^{1,13} \footnotemark[2]$,
  \textbf{Yaying Fei}$^{14} \footnotemark[2]$,\\
  \textbf{Yuheng Cheng}$^{3} \footnotemark[2]$, 
    \textbf{Zhibin Gou}$^{15} \footnotemark[2]$,
    \textbf{Zongze Xu}$^{16} \footnotemark[2]$,
 \textbf{Chenglin Wu}$^{1\S}$
   \vspace{.5em} 
  \\
  $^1$DeepWisdom,
  $^2$Université de Montréal \& Mila,
  $^3$The Chinese University of Hong Kong, Shenzhen, 
   \\
  $^4$Shenzhen University,
  $^5$Fudan University,
  $^6$Renmin University of China, 
   \\
  $^7$Shenzhen Institute of Advanced Technology, Chinese Academy of Sciences, 
  \\
  $^8$King Abdullah University of Science and Technology (KAUST),
  $^9$University of Notre Dame,
  \\
  $^{10}$The University of Hong Kong,
  $^{11}$Yale University, 
  $^{12}$Xiamen University,
  \\
  $^{13}$East China Normal University,
  $^{14}$Beijing University of Technology,
  \\
  $^{15}$Tsinghua University,
  $^{16}$Hohai University
}
  \vspace{-.6cm}

\iclrfinalcopy 
\begin{document}

\maketitle

\renewcommand{\thefootnote}{\S}  
\footnotetext[3]{Bang Liu (E-mail: bang.liu@umontreal.ca) and Chenglin Wu (E-mail: alexanderwu@deepwisdom.ai) are the corresponding authors.}

\begin{abstract}

Large Language Model (LLM)-based agents have shown effectiveness across many applications. However, their use in data science scenarios requiring solving long-term interconnected tasks, dynamic data adjustments and domain expertise remains challenging. 
Previous approaches primarily focus on individual tasks, making it difficult to assess the complete data science workflow. Moreover, they struggle to handle real-time changes in intermediate data and fail to adapt dynamically to evolving task dependencies inherent to data science problems.
In this paper, we present \textbf{Data Interpreter}, an LLM-based agent designed to automatically solve various data science problems end-to-end. 
Our \model{} incorporates two key modules: 1) \textit{Hierarchical Graph Modeling}, which breaks down complex problems into manageable subproblems, enabling dynamic node generation and graph optimization; and 2) \textit{Programmable Node Generation}, a technique that refines and verifies each subproblem to iteratively improve code generation results and robustness.
Extensive experiments consistently demonstrate the superiority of \model{}. 
On InfiAgent-DABench, it achieves a $25\%$ performance boost, raising accuracy from $75.9\%$ to $94.9\%$. For machine learning and open-ended tasks, it improves performance from $88\%$ to $95\%$, and from $60\%$ to $97\%$, respectively.
Moreover, on the MATH dataset, \model{} achieves remarkable performance with a $26\%$ improvement compared to  state-of-the-art baselines.
The code is available at https://github.com/geekan/MetaGPT.

\end{abstract}

\section{Introduction}

Large Language Models (LLMs) have demonstrated remarkable adaptability across a wide range of applications, excelling in areas such as software engineering~\citep{hong2023metagpt}, open-world navigation~\citep{wang2023voyager,wang2023describe,wang2023jarvis,chen2024s}, collaborative intelligence~\citep{zhuge2023mindstorms, zhuge2024language, zhang2024proagent}, and scientific research~\citep{tang2024prioritizing}. However, their performance in data science remains limited.

Data science~\citep{de2022automating,hassan2023chatgpt}, the practice of extracting insights from data, spanning from data gathering to model building and decision-making. 
It integrates multiple disciplines such as computer science, statistics, data visualization, and mathematics~\citep{zhang2023data}. As discussed in~\citep{zhang2024benchmarking,zheng2021evolving}, data science workflows are inherently complex, involving interconnected tasks such as data processing, feature engineering, and model training. 
Solving these tasks requires iterative refinements and real-time adjustments, as both data and requirements continuously evolve.

\begin{figure}[ht]
\centering  
\includegraphics[width=0.87\textwidth]{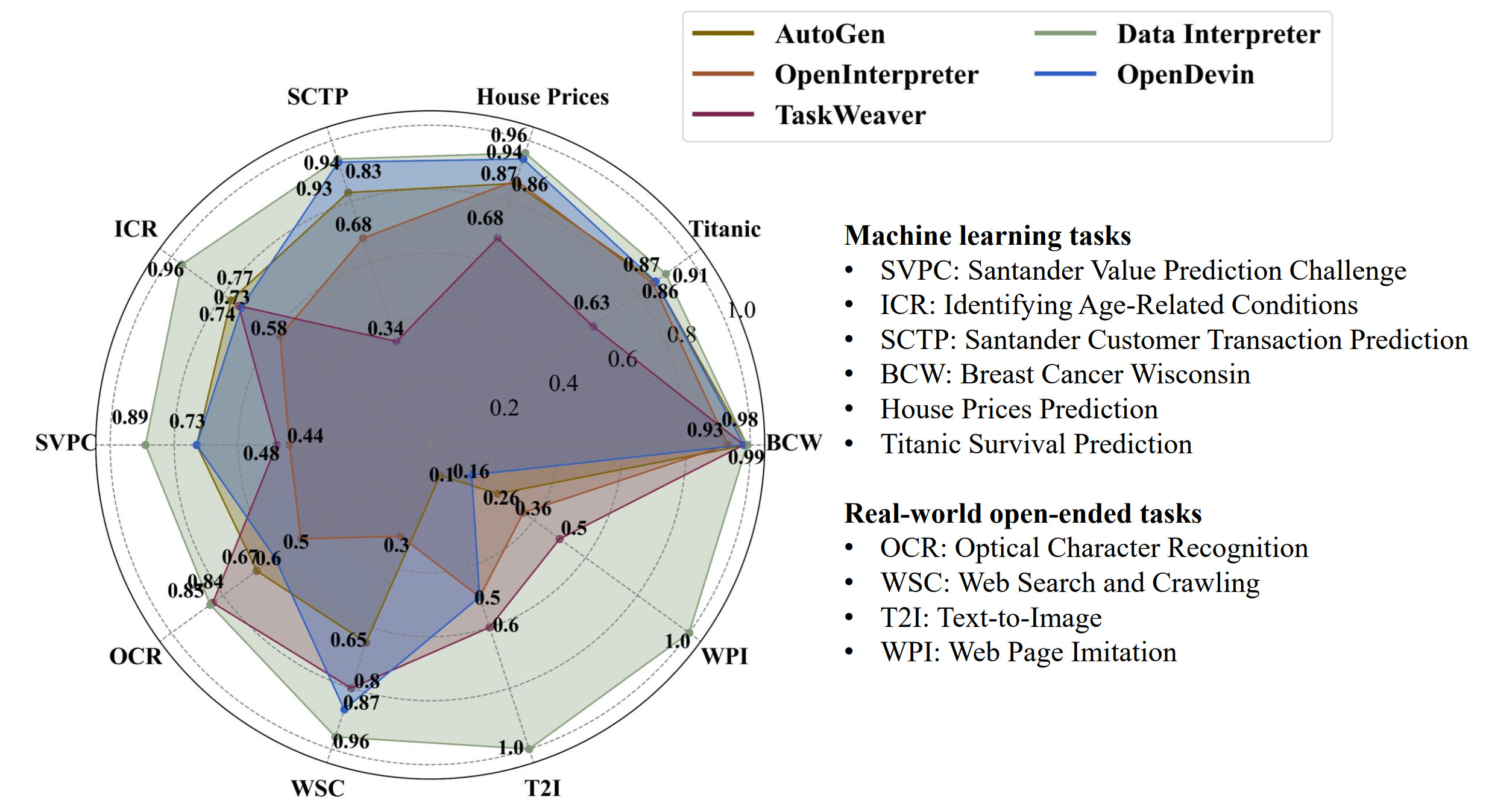}
\caption{Comparison across various open-source frameworks on various data science tasks. We define a unified metric, the \textit{comprehensive score} (\Cref{appendix:exp-metrics}.), to standardize performance evaluation across tasks with different metrics. A higher score indicates better performance. }
\label{img:intro_problem} 
  \vspace{-.3cm}
\end{figure}

Leveraging the extensive knowledge and coding capabilities of LLMs, recent efforts~\citep{shen2024hugginggpt, hollmann2023large, bordt2024data, zhang2024benchmarking,liu2024large} have integrated LLMs into data science tasks. These approaches primarily focus on individual tasks, such as feature engineering~\citep{hollmann2023large}, model selection~\citep{shen2024hugginggpt}, and hyperparameter optimization~\citep{liu2024large}, often operating within fixed pipelines. However, they lack a holistic evaluation of end-to-end workflows, making it difficult to assess the complete data science process. Furthermore, these methods often struggle to handle real-time changes in intermediate data and adapt dynamically to evolving task dependencies. 
While recent works~\citep{wu2023autogen,zhang2023data} have improved performance in data-related tasks, they remain inadequate for machine learning or comprehensive data transformation tasks, involving intricate task interdependencies that require continuous updates and dynamic global planning~\citep{zhang2024benchmarking}.

To address these challenges, we present \textbf{Data Interpreter}, an LLM agent that reframes the data science workflows as a \textit{Hierarchical Graph Modeling} problem, where interconnected tasks are represented as nodes, and their dependencies as edges within the graph. 
This structured representation enables dynamic and flexible task management, allowing the system to adjust to evolving data and task requirements in real-time, and thus efficiently manages the complex, interdependent steps of data science. 
Another core of Data Interpreter is \textit{Programmable Node Generation}, a key innovation that automates the real-time generation, refinement, and verification of nodes in the graph. This ensures that each subproblem is accurately defined and executed, improving the robustness and precision of the workflow. Leveraging the coding capabilities of LLMs, the system dynamically synthesizes and optimizes the graph structure, making it highly adaptable to the demands of complex, evolving data science tasks.

Our experiments demonstrate that Data Interpreter significantly outperforms existing methods across several benchmarks, achieving a 25\% performance boost on the public dataset InfiAgent-DABench, and a 26\% improvement on the MATH dataset. Compared to other open-source frameworks, \model{} consistently shows notable advancements in machine learning and open-ended tasks, as illustrated in \Cref{img:intro_problem}. By rethinking how data science workflows are structured and managed, \model{} sets a new standard for adaptability and efficiency, offering a powerful solution for complex, real-world applications.

\section{Related Work}

\paragraph{LLMs as Data Science Agents}
Large language models (LLMs) have have made significant progress in automating data science tasks, initically focusing on code generation to simplify complex computations involved in reasoning processes~\citep{gao2023pal, chen2022program}. 
Subsequent work introduced code interpreters that leverage function-calling mechanisms, offering greater flexibility in solving complex problems~\citep{zhou2023solving, gou2024tora, wang2024executable}. This interpreter-based approach has now become a mainstream method for enabling LLMs to handle complex reasoning and scientific tasks~\citep{huang2023benchmarking, hassan2023chatgpt, qiao2023taskweaver, zhang2024mlcopilot}.
Recently, \citet{zhang2023data} introduces an LLM-based agent for data analysis, demonstrating capabilities in data processing and exploration within a code-centric framework, but does not evaluate its performance on predictive tasks such as machine learning pipelines.
\citet{guo2024ds} harness LLMs and case-based reasoning to solve data science tasks, leveraging human expertise to enhance the efficiency of LLM-based agents in data science, which is complementary to our work.
~\citet{liu2024large} used LLMs for hyperparameter tuning, focusing on single tasks rather than full pipeline construction and evaluation. 
Therefore, end-to-end frameworks specifically designed for data science tasks remain insufficiently developed.
To address this gap, we propose a unified, general framework specifically designed for data science tasks. Our framework has been rigorously benchmarked across diverse tasks and settings, offering valuable insights into the application and effectiveness of LLMs in data science.

\paragraph{Enhancing LLM with Tools}
Recent research has focused on enhancing LLM capabilities by integrating external tools~\citep{schick2024toolformer, paranjape2023art}. \citet{zhuge2023mindstorms, shen2024hugginggpt} introduced multi-agent systems to tackle multimodal tasks, while \citet{yuan2023craft, liu2023controlllm} proposed frameworks for retrieval and automatic tool selection, eliminating the need to assign tools for specific tasks statically. 
Recent efforts have increasingly focused on integrating tool-using abilities into a structured pipeline, enabling sophisticated task planning, tool invocation~\citep{wu2023visual,shen2024hugginggpt,liang2024taskmatrix}. ~\citet{qian2023creator, yuan2024easytool} discuss the creation and 
instruction of the tool from code-form or lengthy tool documentation to enhance tool utilization efficiency. In this paper, we further advance these ideas by enabling LLMs to dynamic orchestration and combination of multiple tools. Our approach improves practicality by leveraging execution experience, allowing LLMs to select and combine tools as needed independently.

\paragraph{Graph-Based Planning for LLM Agents}
Planning is a critical capability of LLM-based agents, focusing on generating logically structured action or thought roadmaps for specific problems~\citep{huang2024understanding, chen2024s}. Earlier works like CoT~\citep{wei2022chain,yao2022react} decompose complex tasks into subtasks and perform sequential planning. However, due to the complexity of certain problems, a single plan generated by an LLM-based agent is often insufficient. To address this, ToT~\citep{yao2024tree} and GoT~\citep{besta2023graph} introduce automatic tree or graph structures that refine node-level LLM prompts, optimizing connectivity to improve performance. Similarly, DSPy~\citep{khattab2023dspy} abstracts LLM pipelines as text transformation graphs, while PRODIGY~\citep{huang2023prodigy} applies graph-based in-context learning and pre-training methods. Further, \citet{zhuge2024language} enhance node prompts and agent coordination via graph connectivity adjustments, and \citet{vierling2024inputconditionedgraphgeneration} develop a learnable model to dynamically generate edges between agents in a graph, facilitating internal communication.
While these planning approaches excel in various domains, they often struggle with multi-step, task-dependent problems commonly encountered in data science. In this paper, we explore the potential of integrating graph structures with LLM-based agents for data science tasks—an area that remains largely untapped despite emerging related work.

\section{Methodology}
In this section, we first present the foundational formulation of hierarchical graph modeling for data science problems, defining the task graph and action graph in \Cref{subsec:hgm}. Next, we detail the iterative optimization process of the hierarchical graph structure in \Cref{subsec:optimization}. Finally, in \Cref{subsec:png}, we introduce programmable node generation, explaining how we integrate expertise at different granularities to improve the performance of LLMs.

\subsection{Hierarchical Graph Modeling for Complex Task Decomposition}
\label{subsec:hgm}
\begin{figure}[ht]
    \centering
    \includegraphics[width=0.88\textwidth]{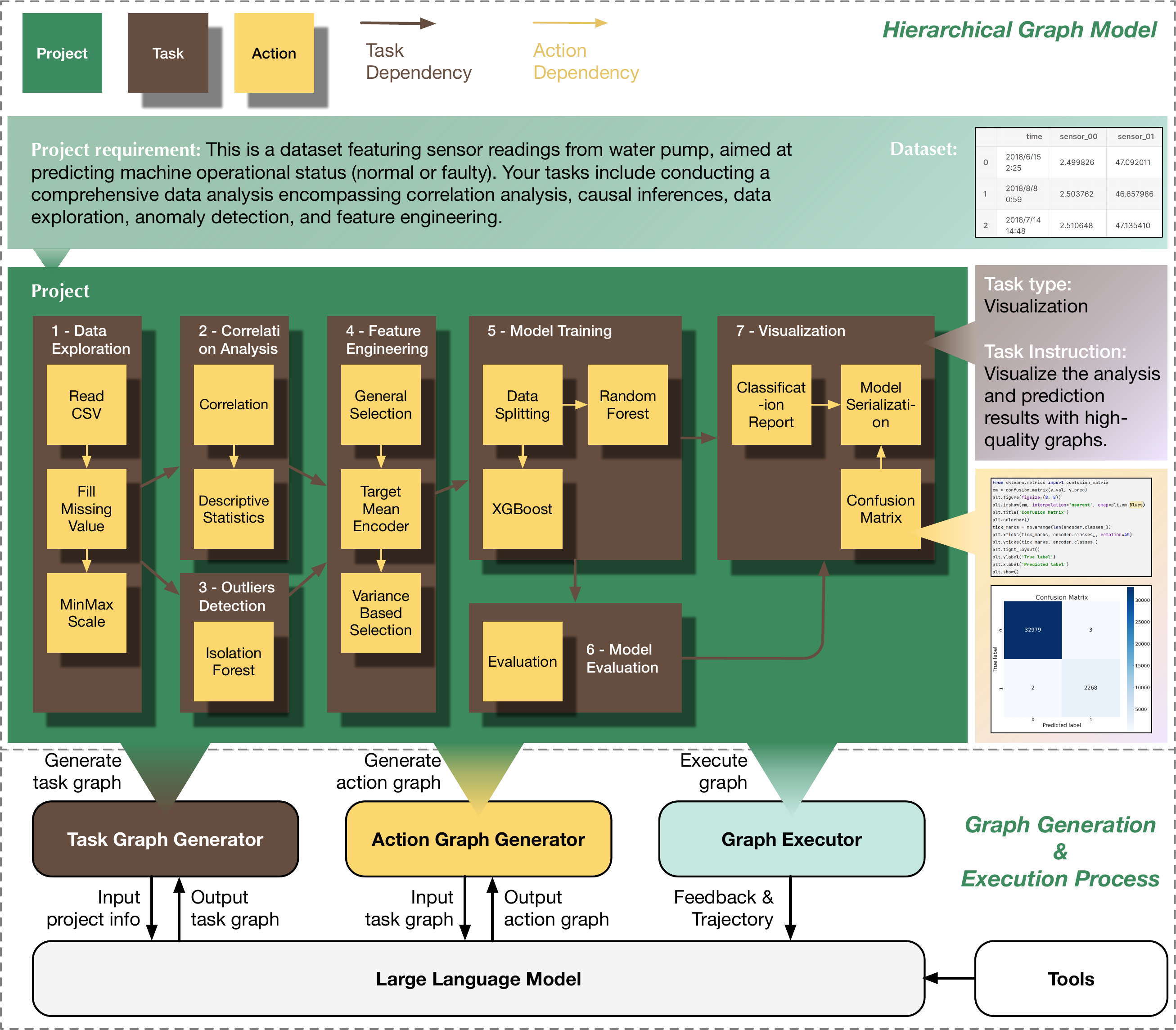}
        \caption{\textbf{Data Interpreter example workflow.} The upper section illustrates how \model{} organizes a data science workflow hierarchically by decomposing project requirements into a task graph, which is then broken down into executable actions. The lower section highlights core modules of \model{}: the \emph{task graph generator}, \emph{action graph generator}, and \emph{graph executor}, which manage task execution and provide real-time feedback. The \emph{graph executor} efficiently executes the action graph using reflection and integrated tools.  
        }
      
    \label{fig:plan-dag}
\end{figure}

Data science problems, particularly those involving machine learning, encompass extensive detailing and long-range workflows, including data pre-processing, feature engineering, and model training. This long-term planning complicates the direct planning of all detailed tasks and coding. 
Drawing inspiration from the application of hierarchical planning in automated machine learning tasks~\citep{mohr2018ml,mubarak2023hierarchical}, we organize the data science workflow via hierarchical structure, which initially decomposes the intricate data science problem into manageable tasks and further break down each task into specific actions executed through code (see \Cref{fig:plan-dag}).

Therefore, solving a data science problem can be formulated as follows: given a task-oriented input \( x \), we seek to apply a series of operators, unified as a function \texttt{P}, to produce an output \( \hat{y} = \texttt{P}(x) \).
Our goal is for \texttt{P} to generate solutions that closely approximate or match the anticipated \(y\). However, due to the complexity of \texttt{P}, which may involve various operations and intermediate data, fully automating the solution to a task is typically challenging.

Leveraging the reasoning ability of LLMs for general task decomposition, our method decomposes the solving process of \texttt{P} into a series of sub-processes \( \texttt{p}_1, \texttt{p}_2, \texttt{p}_3, \ldots \) that can be directly solved and verified.
The primary challenge lies in determining the relationships \( r  = \langle \texttt{p}_i, \texttt{p}_j \rangle \in \mathcal{R} \) between these sub-processes.
Our framework represents all subprocesses as nodes within \texttt{P}, ultimately forming a graph $\mathcal{G}$ that embodies the entire function \texttt{P}:
\begin{equation}
\hat{y} = \mathcal{G}\left( \{\texttt{p}_i(x)\}_{i=1}^{n}, \mathcal{R} \right),
\end{equation}
where \(\mathcal{G}\) represents a Directed Acyclic Graph (DAG) composed of the sub-functions \(\texttt{p}_1, \texttt{p}_2, \texttt{p}_3, \ldots\) interconnected through the relationships \(\mathcal{R}\). This graph illustrates how these sub-functions are combined to generate the final output \(\hat{y}\).
Unlike traditional reinforcement learning (RL) methods for planning~\citep{moerland2023model,schmidhuber2003exploring}, which often require a substantial number of demonstrations to perform domain-specific training, our approach leverages the in-context learning of LLMs. This training-free nature allows our method more adaptable and efficient for general task decomposition.

Improving $\mathcal{R}$ involves achieving an optimal node topology, which has demonstrated robust performance and flexibility in prior research \cite{zhuge2024language}.
In our framework, all subprocesses exchange intermediate results and parameters, represented as \( r = \langle \texttt{p}_i, \texttt{p}_j \rangle \in \mathcal{R} \).
Given the inherent challenges in data science problems \cite{hutter2019automated}, this process can be complex. However, we can optimize the graph topology by refining the relationships between subprocesses. Our objective is:
\begin{equation}
\mathcal{G}^* = \arg \max_{\mathcal{G}} \; \mathbb{E}_{x \sim \mathcal{D}} \left[ \text{Performance}\left( \mathcal{G}\left( \{\texttt{p}_i(x)\}_{i=1}^{n}, \mathcal{R} \right), y \right) \right],
\end{equation}
where \( \mathbb{E}_{x \sim \mathcal{D}} \) denotes the expectation over the data distribution \( \mathcal{D} \), and \( \text{Performance} \) measures the accuracy of the predicted output \( \hat{y} \) against the target output \( y \).
Importantly, within $\mathcal{G}^*$, if a subprocess \(\texttt{p}_i\) proves challenging to solve, it can be further decomposed into smaller, more manageable subprocesses.
Next, we will illustrate the core concepts in our hierarchical graph modeling with an example.

\noindent\textbf{Task Graph.} $\ \ $
\model{} utilizes LLMs to perform task planning, providing only the project requirement as the goal without relying on pre-defined steps or tasks. 
As shown in \Cref{fig:plan-dag}, an example workflow decomposed by \model{} for a machine operational status prediction problem, might include tasks like:\texttt{data exploration}, \texttt{correlation analysis}, \texttt{outliers detection}, \texttt{feature engineering}, \texttt{model training}, \texttt{model evaluation}, and \texttt{visualization}. 
Each task node is defined within the metadata and includes attributes such as task description, task type, status, execution feedback, and dependencies,
collectively form the task-level graph \(\mathcal{G}\), enabling structured task management and execution.
Consequently, during the solving process, the dynamic contextual data are automatically constructed and acquired through the inter-dependencies among tasks, avoiding the need to retrieve the entire context at once while maintaining the relevance of the input context, offering flexibility and scalability for broader data science applications.

\noindent\textbf{Action Graph.} $\ \ $ \label{subsub:action graph}
\model{} breaks down each task into multiple actions using contextual memory, thus forming an action graph. Action graphs can be executed and verified independently, and the synthesis of each action node will be detailed in \Cref{subsec:png}.
As illustrated in~\Cref{fig:plan-dag}, the visualization task is divided into three distinct actions, with the confusion matrix calculation handled by \texttt{sklearn}. The solving process is represented as an action graph, visually captures the relationships between these actions and serves as an implicit representation of the code. Additional runtime examples are provided in~\Cref{fig:graph_examples} in the Appendix.

At a more granular level, action graph iteratively adjusts to handle real-time execution feedback, such as managing failures by refining code or incorporating verification processes, making it a sufficiently granular unit for rapid task adjustments and validation. We explore this optimization process further in~\Cref{subsec:optimization}.

\subsection{Task graph: Iterative Graph Refinement}
\label{subsec:optimization} 

\noindent\textbf{Task Graph Generation and Execution.}
A key advantage of our approach is its ability to dynamically adjust the task graph in response to changing environments, unlike prior methods~\citep{wei2022chain, besta2023graph, yao2022react} and frameworks such as OpenInterpreter~\citep{githubGitHubKillianLucasopeninterpreter} and AutoGen~\citep{wu2023autogen}, which generate static plans for one-time execution. Our method introduces iterative graph optimization, allowing it to adapt to a dynamic environment through continuous updates.

As shown in \Cref{fig:plan-dag}, \model{} uses a task graph generator to initialize the task graph as discussed in \Cref{subsub:action graph}. Each task is then translated into executable code by the action graph generator, which takes into account the outcomes of prior tasks to ensure contextual consistency. The generation process is detailed in \Cref{algo:dynamic_graph_v2}.

To ensure runtime verification and provide real-time feedback during execution, \model{} incorporates a stateful graph executor that manages both execution and debugging using reflection mechanisms~\citep{shinn2024reflexion}. Specifically, if the execution encounters exceptions or fails a verification check, the action graph generator dynamically reflects on the execution results, and then regenerates the code to resolve the issue or optimize the output, providing data-driven feedback. This process is collectively conducted by action graph generator and graph executor.

\begin{algorithm}
\caption{Iterative Graph Execution}
\label{algo:dynamic_graph_v2}
\textbf{Input:} User requirements \( req \), large language model \( LLM \), tool sets \( T \)\\
\textbf{Output:} Optimized graph \( G^* \)
\begin{algorithmic}[1]
\State Set \( M \) as the maximum number of iterations, \( R \) to denote runtime results
\State \( G \gets \text{initialize\_graph}(req, LLM) \) \Comment{Initialize the graph with user requirements}
\While{not \( G.\text{is\_finished}() \)} \Comment{Iterative process until termination condition is met}
    \State \( tn \gets \text{select\_task\_node}(G, LLM) \) 
    \State \( ag \gets \text{initialize\_action\_graph}(tn, T, LLM) \)
    \For{\( i = 1 \) to \( M \)} \Comment{Execute up to M iterations or until success}
        \State \( R \gets \text{execute}(ag) \) 
        \If{\( \text{is\_success}(R) \)}
            \State \textbf{break} \Comment{Exit loop if the action is successful}
        \EndIf
        \State \( ag \gets \text{refine}(tn, R, LLM) \) \Comment{Refine the action graph based on runtime result}
    \EndFor
    \State \( tn \gets \text{update\_node\_state}(tn, ag, R) \) \Comment{Update the state of the task node}
    \State \( G.\text{task\_graph} \gets \text{update\_task\_graph}(G, tn) \) \Comment{Integrate updates into the task graph}
\EndWhile
\State \( G^* \gets \text{finalize\_graph}(G) \) \ \Comment{Save optimized graph}
\State \Return{\( G^* \)}

\end{algorithmic}
\end{algorithm}

\begin{figure}[ht]
    \centering
    \includegraphics[width=0.87\textwidth]
    {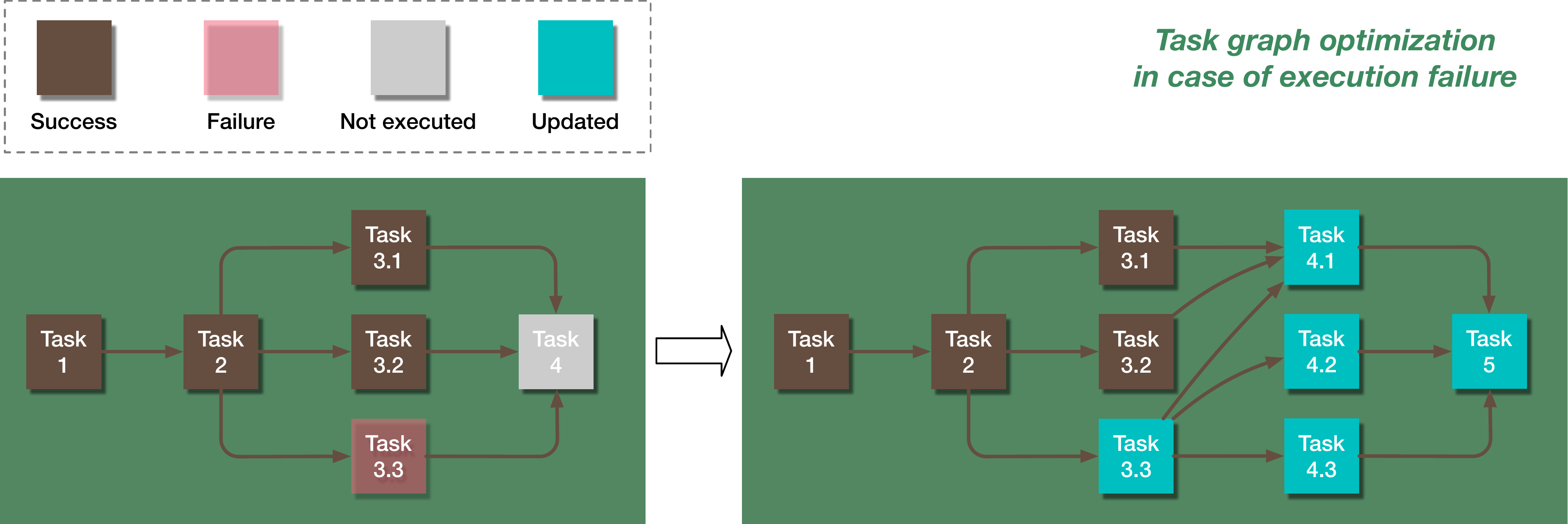}
    \caption{\textbf{Task graph refinement of \model{}.} 
    Task graph refinement for the failed task. After task execution, Task 3.3 fails. The refined task graph integrates existing success tasks, \textcolor{black}{replaces task 3.3 with the updated task 3.3, and introduces new tasks 4.1, 4.2, 4.3 and 5.}
    }
    \label{fig:dynamic_graph_v2}
    \vspace{-0.1in}
\end{figure}

\noindent\textbf{Task Graph Refinement.} $\ \ $
The task graph generator manages tasks, monitors their statuses and dependencies, and dynamically adjusts the task graph by adding, removing, or modifying tasks as needed. Each task is further decomposed into an action graph, which consists of one or several action nodes. Each action graph can be executed and evaluated independently, allowing for granular control and flexibility in the execution process. 
During execution, a task is marked as \texttt{Success} if the corresponding code executes successfully. If execution fails, \model{} leverages LLMs to debug the code based on runtime errors, making up to a predefined number of attempts to resolve the issue. If the problem persists after the set attempts, the task node is flagged as \texttt{Failure}, as shown in \Cref{fig:dynamic_graph_v2}.

For failed tasks, \model{} regenerates the task graph based on current episodic memory and the execution context, as depicted in \Cref{fig:dynamic_graph_v2}. Given the task dependencies, the regenerated task graph is sorted topologically and compared to the original using a prefix matching algorithm~\citep{waldvogel2000fast} to identify differences in task descriptions. This comparison helps identify divergence points (forks), and the final output includes all unchanged tasks before the fork, along with any new or modified tasks after the fork. This approach allows \model{} to efficiently locate the parent node of the failed task and seamlessly integrate the newly generated task and its subsequent tasks into the original graph. It directly leverages the completed memory of all dependent tasks during re-execution, avoiding unnecessary code regeneration or redundant executions.

By employing continuous monitoring and iterative updates, \model{} avoids the inefficiencies associated with generating all tasks upfront. This dynamic adjustment of both the code and planning levels based on task outcomes enables modifications at varying levels of granularity, significantly improving overall efficiency.

\subsection{Action Graph: Programmable Node Generation} \label{subsec:png}
\noindent\textbf{Action Node.} $\ \ $  
An action node, as introduced in \Cref{subsub:action graph}, represents an executable code snippet that encapsulates the computational logic required for task execution. Each action node can encompass data transformations, function calls, or other relevant operations, making it the fundamental unit of execution within the action graph. It integrates both external functions and operators invoked from various tools, as well as non-tool logic derived from libraries such as Pandas and NumPy. By combining tool-based operations and library functions into a single executable code snippet, action nodes ensure uniform and flexible execution across different tasks.

\noindent\textbf{Tool Selection.} $\ \ $  
Effective tool selection and integration, particularly in the context of task-specific requirements, play a crucial role in the success of task execution, as noted in prior research~\citep{qian2023creator, yuan2024easytool, huang2024planning, liu2023controlllm}. In \model{}, we leverage task dependencies to enrich the task-specific context, thereby enhancing the decision-making process for tool selection and code generation.

During the execution of each task \( \texttt{p}_i \in \mathcal{G} \), where \( \mathcal{G} \) represents the task graph, \model{} first retrieves suitable tools before generating the associated code. The task metadata \( q(\texttt{p}_i) \), which includes textual information such as task descriptions and types as well as graph-structured task dependencies, is used as a query to retrieve a list of candidate tools from the available toolset \( T = \{t_1, t_2, \dots, t_n\} \). The model ranks these tools by evaluating their semantic relevance to the task using their functionality schemas \( \mathcal{S}(t_j) \). This produces a ranked list \( R(\texttt{p}_i, T) = \{r_1, r_2, \dots, r_n\} \), where each tool \( t_j \) is ranked according to its suitability for the task.
From this ranked list, \model{} selects the top-\( k \) tools, denoted as \( T_k(\texttt{p}_i) \subseteq T \), to assist in executing task \( \texttt{p}_i \). Importantly, \model{} can bypass tool selection when no suitable tools are found, relying solely on the LLM to generate appropriate code. This flexibility ensures that the system can adapt to a wide range of task requirements without being restricted by tool availability.

\noindent\textbf{Programmable Node Generation.} $\ \ $  
Unlike conventional LLM-based agent frameworks that invoke tools through isolated function calls, \model{} generates comprehensive code snippets that seamlessly integrate selected tools within the broader logic of the task. Based on the tools selected from \( T_k(\texttt{p}_i) \), \model{} dynamically incorporates them into the code, aligning their functionality with the specific task context. This approach allows tools to function in the same manner as standard libraries like NumPy, enabling adaptive tool usage that adjusts to evolving task conditions. For example, in the deployment workflow, the CatCount tool dynamically utilizes its fit and transform functions depending on the task context, as illustrated in \Cref{fig:tool-usage} in the Appendix. 

Our programmable node generation approach not only ensures that tools are used in a context-aware and task-specific manner but also facilitates the seamless integration of domain-specific expertise. By allowing real-time adaptability and optimization of tool usage, \model{} significantly enhances the efficiency and robustness of task execution, representing a novel contribution to LLM-based task automation.

\section{Experiments} \label{main:experiments}
\subsection{Experimental setup}\label{subsec:expsetup}
\noindent\textbf{InfiAgent-DABench:} 
InfiAgent-DABench~\citep{hu2024infiagentdabench} evaluates LLMs in data analysis tasks across 257 challenges from 52 CSV files, covering 7 categories as detailed in \Cref{appendix:exp-datasets}.
We used accuracy as the evaluation metric. \model{} was primarily evaluated with \texttt{gpt-4o} and \texttt{gpt-4-0613} (temperature=0), and compared against XAgent~\citep{xagent2023}, AutoGen~\citep{wu2023autogen}, as well as other baselines reported from~\citep{hu2024infiagentdabench}. 

\noindent\textbf{ML-Benchmark:} 
To evaluate the performance of solving real-world machine learning challenges, We collected 8 datasets from Kaggle for ML-Benchmark (details in \Cref{tab:dataset-info}. We also detailed the evaluation metrics on ML-Benchmark in \Cref{appendix:exp-metrics}. Baselines included XAgent, AutoGen, OpenInterpreter~\citep{githubGitHubKillianLucasopeninterpreter}, TaskWeaver~\citep{qiao2023taskweaver}, and OpenDevin~\citep{wang2024opendevin}. As default, we used \texttt{gpt-4-1106-preview} with temperature set to 0.

\noindent\textbf{Open-ended task benchmark:} 
To verify the capability for dynamic data handling, we also crafted the Open-ended task benchmark comprising 20 tasks. Details about datasets are in the~\Cref{appendix:exp-datasets}. We adopted AutoGen and OpenInterpreter and OpenDevin as baselines with average results reported over three runs. We adopted \texttt{gpt-4-1106-preview} with temperature set to 0.

\noindent\textbf{MATH:} 
We evaluated 4 categories (C.Prob, N.Theory, Prealg, Precalc) of level-5 problems from the MATH dataset~\citep{hendrycks2021measuring}, following the setting of~\citep{wu2023empirical}. Level-5 problems were chosen for their complexity and the challenges in reliable numeric interpretation. We used MathChat~\citep{wu2023empirical} and AutoGen~\citep{wu2023autogen} as baselines for the MATH benchmark.

\subsection{Main result} 
\noindent\textbf{Performance on InfiAgent-DABench.}$\ \ $ \
As demonstrated in \Cref{tab:dabench-score}, with \texttt{gpt-4-0613}, \model{} achieved a score of 73.55, outperforming AutoGen by 2.9\%. Notably, it still did not surpass the performance of directly invoking the LLM. We found this is primarily due to the growing context overhead in the problem-solving process, where the context length exceeds the maximum window size of \texttt{gpt-4-0613}, leading to task failures.
However, by incorporating LLMs like \texttt{gpt-4o} with longer context windows, \model{} demonstrated outstanding performance, improving results by 25\% compared to direct LLM inference. This indicates that \model{} significantly enhances the LLM's multi-step reasoning capabilities across a wide range of data analysis tasks, especially as the number of interaction rounds increases and the context overhead grows.

\begin{table*}[t]
\tablestyle{1.8pt}{1.05}
\caption{
\textbf{Performance comparisons on InfiAgent-DABench.} 
Results marked with an asterisk (*) are reported by~\citet{hu2024infiagentdabench}. Rows marked with a dagger symbol (†) indicate the w/o Agent baseline for comparison. The $\Delta$ column represents the accuracy improvement of the agent framework compared to the w/o agent setups. The best results are highlighted in bold.
}

\renewcommand\tabcolsep{6pt}
\renewcommand\arraystretch{0.9}
\small
\centering
\begin{tabular}{ll|c|c}
\toprule
\textbf{Agent Framework} & \textbf{Model} & \textbf{Accuracy (\%)} & \textbf{$\Delta$ (\%)}\\
\midrule
\multirow{4}{*}{w/o Agent} 
 & \texttt{gemini-pro}            & 56.42* & -\\
& \texttt{gpt-3.5-turbo-0613}  & 60.70* & -\\
 & \texttt{gpt-4-0613}  & 78.99*† & -\\
 & \texttt{gpt-4-0613}  & 75.21 & -\\
&\texttt{gpt-4o}        & 75.92† & - \\
\midrule
XAgent & \texttt{gpt-4-0613}            & 47.53* & -31.46 \\
AutoGen & \texttt{gpt-4-0613}   & 71.49 & -7.50\\
\rowcolor[gray]{.85}
Data Interpreter  &\texttt{gpt-4-0613}   &  73.55 & -5.44 \\
\rowcolor[gray]{.85}
Data Interpreter  & \texttt{gpt-4o}       &  \textbf{94.93}& \textbf{+19.01} \\
\bottomrule
\end{tabular}
\label{tab:dabench-score}
\end{table*}

\begin{table*}[t]
\tablestyle{1.8pt}{1.05}
\caption{
\textbf{Performance comparisons on ML-Benchmark.} 
This table reports the Comprehensive Score of each task. ``WR'', ``BCW'', ``ICR'', ``SCTP'', and ``SVPC''  represent ``Wine recognition'', ``Breast cancer wisconsin'', ``ICR - Identifying age-related conditions'', ``Santander customer transaction prediction'', and ``Santander value prediction challenge'', respectively. 
}
\renewcommand\tabcolsep{5.9pt}
\renewcommand\arraystretch{1.1}
\small
\centering
\begin{tabular}{l|ccccccccccc}
\hline

\hline

\hline

\hline
\multicolumn{1}{l|}{Model / Task} & \multicolumn{1}{c}{WR} & \multicolumn{1}{c}{BCW} & \multicolumn{1}{c}{Titanic}
& \multicolumn{1}{c}{House Prices} & \multicolumn{1}{c}{SCTP} & \multicolumn{1}{c}{ICR} & \multicolumn{1}{c}{SVPC} & \multicolumn{1}{|c}{Avg.} & \multicolumn{1}{|c}{Cost (\$)}
\\
\midrule
\multicolumn{1}{l|}{AutoGen} & \multicolumn{1}{c}{0.96} & \multicolumn{1}{c}{\textbf{0.99}} & \multicolumn{1}{c}{0.87}
& \multicolumn{1}{c}{0.86} & \multicolumn{1}{c}{0.83} & \multicolumn{1}{c}{0.77} & \multicolumn{1}{c}{0.73} & \multicolumn{1}{|c}{0.86} & \multicolumn{1}{|c}{-}
\\

\multicolumn{1}{l|}{OpenInterpreter} & \multicolumn{1}{c}{\textbf{1.00}} & \multicolumn{1}{c}{0.93}
 & \multicolumn{1}{c}{0.86} & \multicolumn{1}{c}{0.87} & \multicolumn{1}{c}{0.68} & \multicolumn{1}{c}{0.58} & \multicolumn{1}{c}{0.44} & \multicolumn{1}{|c}{0.77}  & \multicolumn{1}{|c}{-}
\\

\multicolumn{1}{l|}{TaskWeaver} & \multicolumn{1}{c}{\textbf{1.00}} & \multicolumn{1}{c}{0.98} & \multicolumn{1}{c}{0.63}
& \multicolumn{1}{c}{0.68} & \multicolumn{1}{c}{0.34} & \multicolumn{1}{c}{0.74} & \multicolumn{1}{c}{0.48} & \multicolumn{1}{|c}{0.69} & \multicolumn{1}{|c}{\textbf{0.37}}
\\

\multicolumn{1}{l|}{XAgent} &
\multicolumn{1}{c}{\textbf{1.00}} & \multicolumn{1}{c}{0.97} & \multicolumn{1}{c}{0.42}
& \multicolumn{1}{c}{0.42} & \multicolumn{1}{c}{0} & \multicolumn{1}{c}{0.34} & \multicolumn{1}{c}{0.01} & \multicolumn{1}{|c}{0.45} & \multicolumn{1}{|c}{20.09}
\\

\multicolumn{1}{l|}{OpenDevin} &
\multicolumn{1}{c}{0.98} & \multicolumn{1}{c}{0.98} & \multicolumn{1}{c}{0.87}
& \multicolumn{1}{c}{0.94} & \multicolumn{1}{c}{0.93} & \multicolumn{1}{c}{0.73} & \multicolumn{1}{c}{0.73} & \multicolumn{1}{|c}{0.88}  & \multicolumn{1}{|c}{3.01}
\\

\rowcolor[gray]{.9}
\multicolumn{1}{l|}{\textbf{Data Interpreter}} &
\multicolumn{1}{c}{0.98} & \multicolumn{1}{c}{\textbf{0.99}} & \multicolumn{1}{c}{\textbf{0.91}}
& \multicolumn{1}{c}{\textbf{0.96}} & \multicolumn{1}{c}{\textbf{0.94}} & \multicolumn{1}{c}{\textbf{0.96}} & \multicolumn{1}{c}{\textbf{0.89}} & \multicolumn{1}{|c}{\textbf{0.95}} & \multicolumn{1}{|c}{0.84}
\\
\hline

\hline

\hline

\hline
\end{tabular}
    \label{tab:ml-score}
\end{table*}

\begin{table*}[t]
\tablestyle{1.8pt}{1.05}
\caption{
\textbf{Performance comparisons on Open-ended task benchmark.}
 This table reports the Completion Rate of each task. The tested tasks include ``OCR" (Optical Character Recognition), ``WSC" (Web Search and Crawling), and ``ER" ( Email Reply), ``WPI" (Web Page Imitation), ``IBR" (Image Background Removal), ``T2I" (Text-to-Image), ``I2C" (Image-to-Code) and ``MGG" (Mini Game Generation).
}
\renewcommand\tabcolsep{5.9 pt}
\renewcommand\arraystretch{1.1}
\small
\centering
\begin{tabular}{l|ccccccccccc}
\hline

\hline

\hline

\hline
\multicolumn{1}{l|}{Model / Task} & \multicolumn{1}{c}{OCR} & \multicolumn{1}{c}{WSC} & \multicolumn{1}{c}{ER}
& \multicolumn{1}{c}{WPI} & \multicolumn{1}{c}{IBR} & \multicolumn{1}{c}{T2I} & \multicolumn{1}{c}{I2C} & \multicolumn{1}{c}{MGG} & \multicolumn{1}{|c}{Avg.} & \multicolumn{1}{|c}{Cost (\$)}\\
\midrule
\multicolumn{1}{l|}{AutoGen} & \multicolumn{1}{c}{0.67} & \multicolumn{1}{c}{0.65} & \multicolumn{1}{c}{0.10}
& \multicolumn{1}{c}{0.26} & \multicolumn{1}{c}{\textbf{1.00}} & \multicolumn{1}{c}{0.10} & \multicolumn{1}{c}{0.20} & \multicolumn{1}{c}{0.67} & \multicolumn{1}{|c}{0.46} & \multicolumn{1}{|c}{-}\\

\multicolumn{1}{l|}{OpenInterpreter} & \multicolumn{1}{c}{0.50} & \multicolumn{1}{c}{0.30}
 & \multicolumn{1}{c}{0.10} & \multicolumn{1}{c}{0.36} & \multicolumn{1}{c}{\textbf{1.00}} & \multicolumn{1}{c}{0.50} & \multicolumn{1}{c}{0.25} & \multicolumn{1}{c}{0.20} & \multicolumn{1}{|c}{0.40} & \multicolumn{1}{|c}{-}\\

\multicolumn{1}{l|}{OpenDevin} & \multicolumn{1}{c}{0.60} & \multicolumn{1}{c}{0.87}
 & \multicolumn{1}{c}{0.10} & \multicolumn{1}{c}{0.16} & \multicolumn{1}{c}{\textbf{1.00}} & \multicolumn{1}{c}{0.50} & \multicolumn{1}{c}{0.80} & \multicolumn{1}{c}{0.90} & \multicolumn{1}{|c}{0.60} & \multicolumn{1}{|c}{1.41}\\



\rowcolor[gray]{.9}
\multicolumn{1}{l|}{\textbf{Data Interpreter}} & \multicolumn{1}{c}{\textbf{0.85}} &
 \multicolumn{1}{c}{\textbf{0.96}} & \multicolumn{1}{c}{\textbf{0.98}}
& \multicolumn{1}{c}{\textbf{1.00}} & \multicolumn{1}{c}{\textbf{1.00}} & \multicolumn{1}{c}{\textbf{1.00}} & \multicolumn{1}{c}{\textbf{1.00}} & \multicolumn{1}{c}{\textbf{0.93}} & \multicolumn{1}{|c}{\textbf{0.97}} & \multicolumn{1}{|c}{\textbf{0.41}}\\

\hline

\hline

\hline

\hline
\end{tabular}
    \label{tab:open-ended-task-score}
\end{table*}

\noindent\textbf{Performance on ML-Benchmark.}$\ \ $ \
As shown in \Cref{tab:ml-score}, \model{} achieved a comprehensive score of 0.95 across tasks, outperforming AutoGen (0.86) and OpenDevin (0.88) by 10.3\% and 7.9\%, respectively. It was the only framework to achieve a score above 0.9 on tasks such as Titanic, House Prices, SCTP, and ICR. Additionally, the Data Interpreter demonstrated a significant advantage over other frameworks, with improvements of 31.5\% and 21.9\% over OpenDevin on the ICR and SVPC tasks, respectively. Notably, \model{} solved the tasks more efficiently, achieving an average score of \$ 0.84 while operating at only 27.9\% of OpenDevin's cost. \model{} consistently completed all mandatory processes across datasets, maintaining superior performance. Further details can be found in \Cref{tab:full-ml-score} in the Appendix.

\noindent\textbf{Performance on Open-ended tasks.}$\ \ $ \
~\Cref{tab:open-ended-task-score} illustrates that the Data Interpreter achieved a completion rate of 0.97, marking a substantial 110.8\% improvement compared to AutoGen and 61.7\% improvement compared to OpenDevin. 
In OCR-related tasks, the Data Interpreter maintained an average completion rate of 0.85, outperforming AutoGen, OpenInterpreter OpenDevin by 26.8\%, 70.0\% and 41.7\%, respectively. 
In tasks requiring multiple steps and utilizing multimodal tools/interfaces, such as WPI, I2C, and T2I, the Data Interpreter emerged as the sole method to execute all steps. 
Baseline frameworks failed to log in and obtain the status for the ER task, resulting in a lower completion rate. In contrast, \model{} dynamically adjusted to task requirements, achieving a completion rate of 0.97.

\begin{figure*}[ht]
    \centering
    \includegraphics[width=0.9\textwidth]{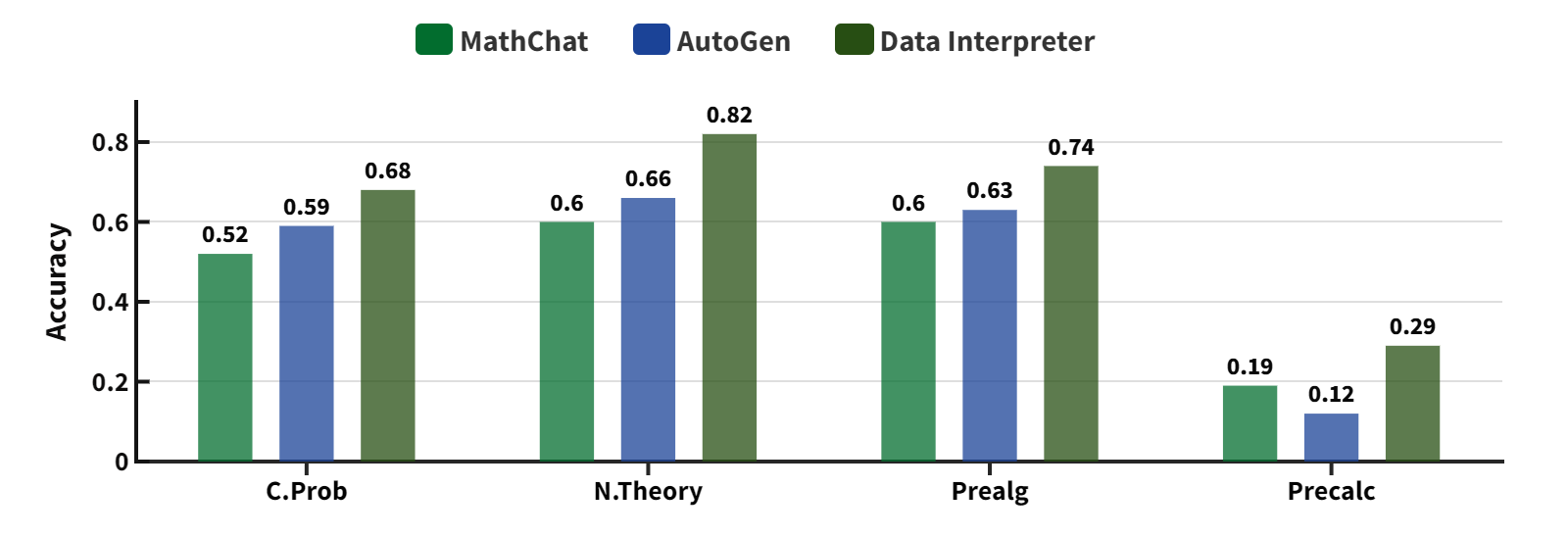}
    \caption{\textbf{Performance on the MATH dataset.} We evaluate all the problems with difficulty level 5 from 4 categories of the MATH dataset. 
    }
    \label{fig:math result}
     \vspace{-0.1in}
\end{figure*}

\noindent\textbf{Performance on math problem.}$\ \ $ \
As illustrated in the \Cref{fig:math result}, \model{} achieved the best results across all tested categories, reaching 0.82 accuracy in the N.Theory category, marking a 0.16 improvement over the performance of AutoGen. 
In the most challenging category, Precalc, \model{} obtained an accuracy of 0.29, an increase of 0.17 compared to AutoGen. 
On average, our Data Interpreter showed 26.5\% relative improvement compared to AutoGen. 

\begin{table*}[ht]
\tablestyle{1.8pt}{1.05}
\caption{
\textbf{Ablation on core modules.} Evaluated with Comprehensive Score on ML-Benchmark. ``IGR" stands for Iterative Graph Refinement, and ``PNG" denotes Programmable Node Generation. ``ICR", ``SCTP", and ``SVPC" represent ``ICR - Identifying age-related conditions", ``Santander customer transaction prediction", and ``Santander value prediction challenge", respectively. }

\renewcommand\tabcolsep{5.9pt}
\renewcommand\arraystretch{1.2}

\small
\centering
\begin{tabular}{cccc|cccc}
\hline

\hline

\hline

\hline
\multicolumn{1}{c}{Code execution} & \multicolumn{1}{c}{IGR} &\multicolumn{1}{c|}{PNG} & \multicolumn{1}{c}{House Prices} & \multicolumn{1}{c}{SCTP} & \multicolumn{1}{c}{SVPC} &\multicolumn{1}{c}{ICR} & \multicolumn{1}{|c}{Avg.} \\
\midrule
\multicolumn{1}{c}{\checkmark} &\multicolumn{1}{c}{} & \multicolumn{1}{c|}{} & \multicolumn{1}{c}{0.51} & \multicolumn{1}{c}{0.17} & \multicolumn{1}{c}{0.66} & \multicolumn{1}{c}{0.17} & \multicolumn{1}{|c}{0.37} \\
\multicolumn{1}{c}{\checkmark} &\multicolumn{1}{c}{\checkmark} & \multicolumn{1}{c|}{} & \multicolumn{1}{c}{0.96} & \multicolumn{1}{c}{0.91} & \multicolumn{1}{c}{0.80} & \multicolumn{1}{c}{0.74} & \multicolumn{1}{|c}{0.85} \\

\rowcolor[gray]{.9}
\multicolumn{1}{c}{\checkmark} &\multicolumn{1}{c}{\checkmark} & \multicolumn{1}{c|}{\checkmark} & \multicolumn{1}{c}{\textbf{0.96}} & \multicolumn{1}{c}{\textbf{0.95}} & \multicolumn{1}{c}{\textbf{0.89}} & \multicolumn{1}{c}{\textbf{0.96}} & \multicolumn{1}{|c}{\textbf{0.94}} \\
\hline

\hline

\hline

\hline
\end{tabular}
\label{tab:cs-ml-score-abl}
\end{table*}

\subsection{Ablation study}
\noindent\textbf{Ablation on core modules.}$\ \ $ \
We conducted ablation experiments with three configurations on the ML-Benchmark. First, we used ReAct~\citep{yao2022react} for code execution with simplified prompts, followed by the addition of iterative graph refinement, and finally, programmable node generation was introduced, using the Data Interpreter as the default. As shown in \Cref{tab:cs-ml-score-abl}, iterative graph refinement improved performance by 0.48, enhancing dataset preparation and real-time tracking. Programmable node generation further boosted the comprehensive score by 10.6\%, reaching 0.94.
We detailed the results in \Cref{tab:full-ml-score-abl}.

\begin{figure}[th]
    \centering
    \includegraphics[width=0.8\textwidth]{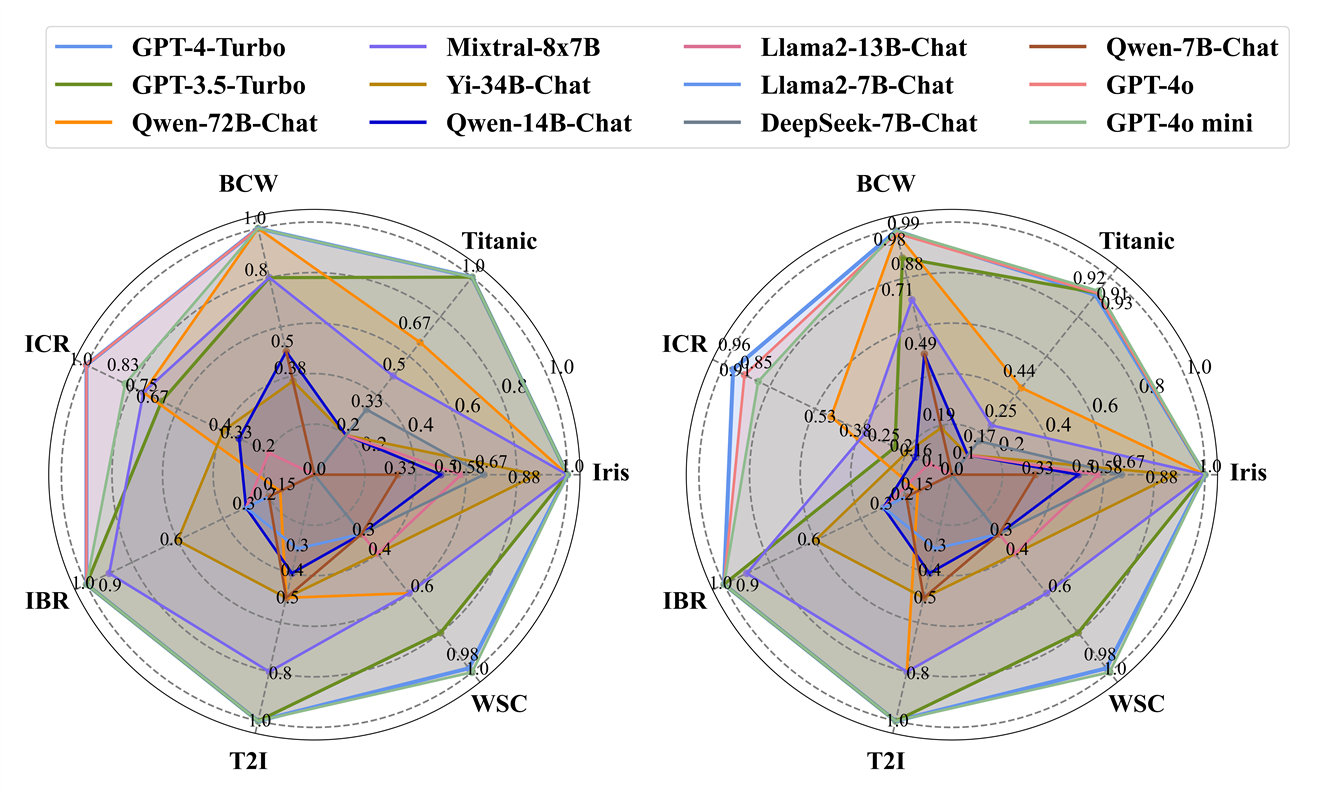}
    \caption{\textbf{Evaluation on ML-Benchmark with different LLMs.} Left: completion rate. Right: comprehensive score. }
    \label{fig: different llms}
    \vspace{-0.1in}
\end{figure}

\noindent\textbf{Ablation on different base LLMs.}$\ \ $ \
Based on GPT-4o and GPT-4o-mini, \model{} shows further improvement in task completion across a wide range of tasks, as illustrated in~\Cref{fig: different llms}.
In machine learning tasks, LLMs like Qwen-72B-Chat~\citep{bai2023qwen} and Mixtral-8x7B~\citep{jiang2024mixtral} performed comparably to GPT-3.5-Turbo, while smaller LLMs experienced performance degradation. Our Data Interpreter handled data loading and analysis effectively with smaller models but had limitations with tasks requiring advanced coding proficiency. Smaller LLMs also encountered failures when using tools to handle images or webpage results, as shown in~\Cref{fig: different llms}.

\section{Conclusion}
In this work, we present the Data Interpreter, an LLM-based agent designed to tackle data science challenges via hierarchical graph representation. Our framework continuously monitors data changes and adapts to dynamic environments through iterative task refinement and graph optimization. It enhances data analysis and machine learning performance, and improves reasoning capabilities through hierarchical decomposition, fine-grained execution, validation, and iterative modifications. Combined with the LLM's planning and coding abilities, this approach effectively solves tasks requiring complex multi-step reasoning.
Extensive evaluations demonstrate that our Data Interpreter outperforms various open-source frameworks in machine learning tasks, mathematical problems, and real-world applications, marking a significant advancement in the capabilities of LLM-based agents for data science.

\clearpage

\bibliography{iclr2025_conference}

\begin{thebibliography}{61}
\providecommand{\natexlab}[1]{#1}
\providecommand{\url}[1]{\texttt{#1}}
\expandafter\ifx\csname urlstyle\endcsname\relax
  \providecommand{\doi}[1]{doi: #1}\else
  \providecommand{\doi}{doi: \begingroup \urlstyle{rm}\Url}\fi

\bibitem[Bai et~al.(2023)Bai, Bai, Chu, Cui, Dang, Deng, Fan, Ge, Han, Huang, et~al.]{bai2023qwen}
Jinze Bai, Shuai Bai, Yunfei Chu, Zeyu Cui, Kai Dang, Xiaodong Deng, Yang Fan, Wenbin Ge, Yu~Han, Fei Huang, et~al.
\newblock Qwen technical report, 2023.

\bibitem[Besta et~al.(2023)Besta, Blach, Kubicek, Gerstenberger, Gianinazzi, Gajda, Lehmann, Podstawski, Niewiadomski, Nyczyk, et~al.]{besta2023graph}
Maciej Besta, Nils Blach, Ales Kubicek, Robert Gerstenberger, Lukas Gianinazzi, Joanna Gajda, Tomasz Lehmann, Michal Podstawski, Hubert Niewiadomski, Piotr Nyczyk, et~al.
\newblock Graph of thoughts: Solving elaborate problems with large language models.
\newblock \emph{arXiv preprint}, 2023.

\bibitem[Bordt et~al.(2024)Bordt, Lengerich, Nori, and Caruana]{bordt2024data}
Sebastian Bordt, Ben Lengerich, Harsha Nori, and Rich Caruana.
\newblock Data science with llms and interpretable models, 2024.

\bibitem[Chen et~al.(2024)Chen, Jiang, Lu, and Zhang]{chen2024s}
Jiaqi Chen, Yuxian Jiang, Jiachen Lu, and Li~Zhang.
\newblock S-agents: self-organizing agents in open-ended environment, 2024.

\bibitem[Chen et~al.(2022)Chen, Ma, Wang, and Cohen]{chen2022program}
Wenhu Chen, Xueguang Ma, Xinyi Wang, and William~W Cohen.
\newblock Program of thoughts prompting: Disentangling computation from reasoning for numerical reasoning tasks, 2022.

\bibitem[De~Bie et~al.(2022)De~Bie, De~Raedt, Hern{\'a}ndez-Orallo, Hoos, Smyth, and Williams]{de2022automating}
Tijl De~Bie, Luc De~Raedt, Jos{\'e} Hern{\'a}ndez-Orallo, Holger~H Hoos, Padhraic Smyth, and Christopher~KI Williams.
\newblock Automating data science.
\newblock \emph{Communications of the ACM}, 65\penalty0 (3):\penalty0 76--87, 2022.

\bibitem[Gao et~al.(2023)Gao, Madaan, Zhou, Alon, Liu, Yang, Callan, and Neubig]{gao2023pal}
Luyu Gao, Aman Madaan, Shuyan Zhou, Uri Alon, Pengfei Liu, Yiming Yang, Jamie Callan, and Graham Neubig.
\newblock Pal: Program-aided language models.
\newblock In \emph{ICML}, 2023.

\bibitem[Gou et~al.(2024)Gou, Shao, Gong, yelong shen, Yang, Huang, Duan, and Chen]{gou2024tora}
Zhibin Gou, Zhihong Shao, Yeyun Gong, yelong shen, Yujiu Yang, Minlie Huang, Nan Duan, and Weizhu Chen.
\newblock To{RA}: A tool-integrated reasoning agent for mathematical problem solving.
\newblock In \emph{The Twelfth International Conference on Learning Representations}, 2024.
\newblock URL \url{https://openreview.net/forum?id=Ep0TtjVoap}.

\bibitem[Guo et~al.(2024)Guo, Deng, Wen, Chen, Chang, and Wang]{guo2024ds}
Siyuan Guo, Cheng Deng, Ying Wen, Hechang Chen, Yi~Chang, and Jun Wang.
\newblock Ds-agent: Automated data science by empowering large language models with case-based reasoning.
\newblock \emph{arXiv preprint arXiv:2402.17453}, 2024.

\bibitem[Hassan et~al.(2023)Hassan, Knipper, and Santu]{hassan2023chatgpt}
Md~Mahadi Hassan, Alex Knipper, and Shubhra Kanti~Karmaker Santu.
\newblock Chatgpt as your personal data scientist, 2023.

\bibitem[Hendrycks et~al.(2021)Hendrycks, Burns, Kadavath, Arora, Basart, Tang, Song, and Steinhardt]{hendrycks2021measuring}
Dan Hendrycks, Collin Burns, Saurav Kadavath, Akul Arora, Steven Basart, Eric Tang, Dawn Song, and Jacob Steinhardt.
\newblock Measuring mathematical problem solving with the math dataset, 2021.

\bibitem[Hollmann et~al.(2023)Hollmann, Müller, and Hutter]{hollmann2023large}
Noah Hollmann, Samuel Müller, and Frank Hutter.
\newblock Large language models for automated data science: Introducing caafe for context-aware automated feature engineering, 2023.

\bibitem[Hong et~al.(2023)Hong, Zhuge, Chen, Zheng, Cheng, Wang, Zhang, Wang, Yau, Lin, et~al.]{hong2023metagpt}
Sirui Hong, Mingchen Zhuge, Jonathan Chen, Xiawu Zheng, Yuheng Cheng, Jinlin Wang, Ceyao Zhang, Zili Wang, Steven Ka~Shing Yau, Zijuan Lin, et~al.
\newblock Metagpt: Meta programming for multi-agent collaborative framework.
\newblock In \emph{The Twelfth International Conference on Learning Representations}, 2023.

\bibitem[Hu et~al.(2024)Hu, Zhao, Wei, Chai, Ma, Wang, Wang, Su, Xu, Zhu, Cheng, Yuan, Li, Kuang, Yang, Yang, and Wu]{hu2024infiagentdabench}
Xueyu Hu, Ziyu Zhao, Shuang Wei, Ziwei Chai, Qianli Ma, Guoyin Wang, Xuwu Wang, Jing Su, Jingjing Xu, Ming Zhu, Yao Cheng, Jianbo Yuan, Jiwei Li, Kun Kuang, Yang Yang, Hongxia Yang, and Fei Wu.
\newblock Infiagent-dabench: Evaluating agents on data analysis tasks, 2024.

\bibitem[Huang et~al.(2023{\natexlab{a}})Huang, Ren, Chen, Kržmanc, Zeng, Liang, and Leskovec]{huang2023prodigy}
Qian Huang, Hongyu Ren, Peng Chen, Gregor Kržmanc, Daniel Zeng, Percy Liang, and Jure Leskovec.
\newblock Prodigy: Enabling in-context learning over graphs, 2023{\natexlab{a}}.

\bibitem[Huang et~al.(2023{\natexlab{b}})Huang, Vora, Liang, and Leskovec]{huang2023benchmarking}
Qian Huang, Jian Vora, Percy Liang, and Jure Leskovec.
\newblock Benchmarking large language models as ai research agents, 2023{\natexlab{b}}.

\bibitem[Huang et~al.(2024{\natexlab{a}})Huang, Zhong, Lu, Zhu, Gao, Liu, Hou, Zeng, Wang, Shang, et~al.]{huang2024planning}
Shijue Huang, Wanjun Zhong, Jianqiao Lu, Qi~Zhu, Jiahui Gao, Weiwen Liu, Yutai Hou, Xingshan Zeng, Yasheng Wang, Lifeng Shang, et~al.
\newblock Planning, creation, usage: Benchmarking llms for comprehensive tool utilization in real-world complex scenarios, 2024{\natexlab{a}}.

\bibitem[Huang et~al.(2024{\natexlab{b}})Huang, Liu, Chen, Wang, Wang, Lian, Wang, Tang, and Chen]{huang2024understanding}
Xu~Huang, Weiwen Liu, Xiaolong Chen, Xingmei Wang, Hao Wang, Defu Lian, Yasheng Wang, Ruiming Tang, and Enhong Chen.
\newblock Understanding the planning of llm agents: A survey, 2024{\natexlab{b}}.

\bibitem[Hutter et~al.(2019)Hutter, Kotthoff, and Vanschoren]{hutter2019automated}
Frank Hutter, Lars Kotthoff, and Joaquin Vanschoren.
\newblock \emph{Automated machine learning: methods, systems, challenges}.
\newblock Springer Nature, 2019.

\bibitem[Jiang et~al.(2024)Jiang, Sablayrolles, Roux, Mensch, Savary, Bamford, Chaplot, Casas, Hanna, Bressand, et~al.]{jiang2024mixtral}
Albert~Q Jiang, Alexandre Sablayrolles, Antoine Roux, Arthur Mensch, Blanche Savary, Chris Bamford, Devendra~Singh Chaplot, Diego de~las Casas, Emma~Bou Hanna, Florian Bressand, et~al.
\newblock Mixtral of experts, 2024.

\bibitem[Khattab et~al.(2023)Khattab, Singhvi, Maheshwari, Zhang, Santhanam, Vardhamanan, Haq, Sharma, Joshi, Moazam, Miller, Zaharia, and Potts]{khattab2023dspy}
Omar Khattab, Arnav Singhvi, Paridhi Maheshwari, Zhiyuan Zhang, Keshav Santhanam, Sri Vardhamanan, Saiful Haq, Ashutosh Sharma, Thomas~T. Joshi, Hanna Moazam, Heather Miller, Matei Zaharia, and Christopher Potts.
\newblock Dspy: Compiling declarative language model calls into self-improving pipelines, 2023.

\bibitem[Liang et~al.(2024)Liang, Wu, Song, Wu, Xia, Liu, Ou, Lu, Ji, Mao, et~al.]{liang2024taskmatrix}
Yaobo Liang, Chenfei Wu, Ting Song, Wenshan Wu, Yan Xia, Yu~Liu, Yang Ou, Shuai Lu, Lei Ji, Shaoguang Mao, et~al.
\newblock Taskmatrix. ai: Completing tasks by connecting foundation models with millions of apis.
\newblock \emph{Intelligent Computing}, 3:\penalty0 0063, 2024.

\bibitem[Liu et~al.(2024)Liu, Gao, and Li]{liu2024large}
Siyi Liu, Chen Gao, and Yong Li.
\newblock Large language model agent for hyper-parameter optimization.
\newblock \emph{arXiv preprint arXiv:2402.01881}, 2024.

\bibitem[Liu et~al.(2023)Liu, Lai, Gao, Cui, Li, Zhu, Lu, Chen, Qiao, Dai, et~al.]{liu2023controlllm}
Zhaoyang Liu, Zeqiang Lai, Zhangwei Gao, Erfei Cui, Zhiheng Li, Xizhou Zhu, Lewei Lu, Qifeng Chen, Yu~Qiao, Jifeng Dai, et~al.
\newblock Controlllm: Augment language models with tools by searching on graphs, 2023.

\bibitem[Liu et~al.(2021)Liu, Pavao, Xu, Escalera, Ferreira, Guyon, Hong, Hutter, Ji, Junior, et~al.]{liu2021winning}
Zhengying Liu, Adrien Pavao, Zhen Xu, Sergio Escalera, Fabio Ferreira, Isabelle Guyon, Sirui Hong, Frank Hutter, Rongrong Ji, Julio CS~Jacques Junior, et~al.
\newblock Winning solutions and post-challenge analyses of the chalearn autodl challenge 2019.
\newblock \emph{TPAMI}, 2021.

\bibitem[Lucas(2023)]{githubGitHubKillianLucasopeninterpreter}
Killian Lucas.
\newblock {G}it{H}ub - {K}illian{L}ucas/open-interpreter: {A} natural language interface for computers --- github.com.
\newblock \url{https://github.com/KillianLucas/open-interpreter}, 2023.

\bibitem[Moerland et~al.(2023)Moerland, Broekens, Plaat, Jonker, et~al.]{moerland2023model}
Thomas~M Moerland, Joost Broekens, Aske Plaat, Catholijn~M Jonker, et~al.
\newblock Model-based reinforcement learning: A survey.
\newblock \emph{Foundations and Trends{\textregistered} in Machine Learning}, 16\penalty0 (1):\penalty0 1--118, 2023.

\bibitem[Mohr et~al.(2018)Mohr, Wever, and H{\"u}llermeier]{mohr2018ml}
Felix Mohr, Marcel Wever, and Eyke H{\"u}llermeier.
\newblock Ml-plan: Automated machine learning via hierarchical planning.
\newblock \emph{Machine Learning}, 2018.

\bibitem[Mubarak \& Koeshidayatullah(2023)Mubarak and Koeshidayatullah]{mubarak2023hierarchical}
Yousef Mubarak and Ardiansyah Koeshidayatullah.
\newblock Hierarchical automated machine learning (automl) for advanced unconventional reservoir characterization.
\newblock \emph{Scientific Reports}, 2023.

\bibitem[Paranjape et~al.(2023)Paranjape, Lundberg, Singh, Hajishirzi, Zettlemoyer, and Ribeiro]{paranjape2023art}
Bhargavi Paranjape, Scott Lundberg, Sameer Singh, Hannaneh Hajishirzi, Luke Zettlemoyer, and Marco~Tulio Ribeiro.
\newblock Art: Automatic multi-step reasoning and tool-use for large language models, 2023.

\bibitem[Qian et~al.(2023)Qian, Han, Fung, Qin, Liu, and Ji]{qian2023creator}
Cheng Qian, Chi Han, Yi~Fung, Yujia Qin, Zhiyuan Liu, and Heng Ji.
\newblock Creator: Tool creation for disentangling abstract and concrete reasoning of large language models.
\newblock In \emph{Findings of EMNLP}, 2023.

\bibitem[Qiao et~al.(2023)Qiao, Li, Zhang, He, Kang, Zhang, Yang, Dong, Zhang, Wang, Ma, Zhao, Qin, Qin, Du, Xu, Lin, Rajmohan, and Zhang]{qiao2023taskweaver}
Bo~Qiao, Liqun Li, Xu~Zhang, Shilin He, Yu~Kang, Chaoyun Zhang, Fangkai Yang, Hang Dong, Jue Zhang, Lu~Wang, Minghua Ma, Pu~Zhao, Si~Qin, Xiaoting Qin, Chao Du, Yong Xu, Qingwei Lin, Saravan Rajmohan, and Dongmei Zhang.
\newblock Taskweaver: A code-first agent framework, 2023.

\bibitem[Schick et~al.(2024)Schick, Dwivedi-Yu, Dess{\`\i}, Raileanu, Lomeli, Hambro, Zettlemoyer, Cancedda, and Scialom]{schick2024toolformer}
Timo Schick, Jane Dwivedi-Yu, Roberto Dess{\`\i}, Roberta Raileanu, Maria Lomeli, Eric Hambro, Luke Zettlemoyer, Nicola Cancedda, and Thomas Scialom.
\newblock Toolformer: Language models can teach themselves to use tools.
\newblock \emph{NeurIPS}, 2024.

\bibitem[Schmidhuber(2003)]{schmidhuber2003exploring}
Juergen Schmidhuber.
\newblock Exploring the predictable.
\newblock In \emph{Advances in evolutionary computing: theory and applications}, pp.\  579--612. Springer, 2003.

\bibitem[Shen et~al.(2024)Shen, Song, Tan, Li, Lu, and Zhuang]{shen2024hugginggpt}
Yongliang Shen, Kaitao Song, Xu~Tan, Dongsheng Li, Weiming Lu, and Yueting Zhuang.
\newblock Hugginggpt: Solving ai tasks with chatgpt and its friends in hugging face.
\newblock \emph{NeurIPS}, 2024.

\bibitem[Shinn et~al.(2024)Shinn, Cassano, Gopinath, Narasimhan, and Yao]{shinn2024reflexion}
Noah Shinn, Federico Cassano, Ashwin Gopinath, Karthik Narasimhan, and Shunyu Yao.
\newblock Reflexion: Language agents with verbal reinforcement learning, 2024.

\bibitem[Tang et~al.(2024)Tang, Jin, Zhu, Yuan, Zhang, Zhou, Qu, Zhao, Tang, Zhang, et~al.]{tang2024prioritizing}
Xiangru Tang, Qiao Jin, Kunlun Zhu, Tongxin Yuan, Yichi Zhang, Wangchunshu Zhou, Meng Qu, Yilun Zhao, Jian Tang, Zhuosheng Zhang, et~al.
\newblock Prioritizing safeguarding over autonomy: Risks of llm agents for science, 2024.

\bibitem[Team(2023)]{xagent2023}
XAgent Team.
\newblock Xagent: An autonomous agent for complex task solving.
\newblock \url{https://github.com/OpenBMB/XAgent}, 2023.

\bibitem[Vierling et~al.(2024)Vierling, Fu, and Chen]{vierling2024inputconditionedgraphgeneration}
Lukas Vierling, Jie Fu, and Kai Chen.
\newblock Input conditioned graph generation for language agents, 2024.
\newblock URL \url{https://arxiv.org/abs/2406.11555}.

\bibitem[Waldvogel(2000)]{waldvogel2000fast}
Marcel Waldvogel.
\newblock Fast longest prefix matching: algorithms, analysis, and applications.
\newblock \emph{Doctoral dissertation, SWISS FEDERAL INSTITUTE OF TECHNOLOGY ZURICH}, 2000.

\bibitem[Wang et~al.(2023{\natexlab{a}})Wang, Xie, Jiang, Mandlekar, Xiao, Zhu, Fan, and Anandkumar]{wang2023voyager}
Guanzhi Wang, Yuqi Xie, Yunfan Jiang, Ajay Mandlekar, Chaowei Xiao, Yuke Zhu, Linxi Fan, and Anima Anandkumar.
\newblock Voyager: An open-ended embodied agent with large language models.
\newblock \emph{arXiv preprint arXiv:2305.16291}, 2023{\natexlab{a}}.

\bibitem[Wang et~al.(2024{\natexlab{a}})Wang, Chen, Yuan, Zhang, Li, Peng, and Ji]{wang2024executable}
Xingyao Wang, Yangyi Chen, Lifan Yuan, Yizhe Zhang, Yunzhu Li, Hao Peng, and Heng Ji.
\newblock Executable code actions elicit better llm agents, 2024{\natexlab{a}}.

\bibitem[Wang et~al.(2024{\natexlab{b}})Wang, Li, Song, Xu, Tang, Zhuge, Pan, Song, Li, Singh, et~al.]{wang2024opendevin}
Xingyao Wang, Boxuan Li, Yufan Song, Frank~F Xu, Xiangru Tang, Mingchen Zhuge, Jiayi Pan, Yueqi Song, Bowen Li, Jaskirat Singh, et~al.
\newblock Opendevin: An open platform for ai software developers as generalist agents.
\newblock \emph{arXiv preprint arXiv:2407.16741}, 2024{\natexlab{b}}.

\bibitem[Wang et~al.(2023{\natexlab{b}})Wang, Cai, Chen, Liu, Ma, and Liang]{wang2023describe}
Zihao Wang, Shaofei Cai, Guanzhou Chen, Anji Liu, Xiaojian Ma, and Yitao Liang.
\newblock Describe, explain, plan and select: Interactive planning with large language models enables open-world multi-task agents.
\newblock In \emph{NeurIPS}, 2023{\natexlab{b}}.

\bibitem[Wang et~al.(2023{\natexlab{c}})Wang, Cai, Liu, Jin, Hou, Zhang, Lin, He, Zheng, Yang, Ma, and Liang]{wang2023jarvis}
Zihao Wang, Shaofei Cai, Anji Liu, Yonggang Jin, Jinbing Hou, Bowei Zhang, Haowei Lin, Zhaofeng He, Zilong Zheng, Yaodong Yang, Xiaojian Ma, and Yitao Liang.
\newblock Jarvis-1: Open-world multi-task agents with memory-augmented multimodal language models.
\newblock \emph{arXiv preprint arXiv:2311.05997}, 2023{\natexlab{c}}.

\bibitem[Wei et~al.(2022)Wei, Wang, Schuurmans, Bosma, Xia, Chi, Le, Zhou, et~al.]{wei2022chain}
Jason Wei, Xuezhi Wang, Dale Schuurmans, Maarten Bosma, Fei Xia, Ed~Chi, Quoc~V Le, Denny Zhou, et~al.
\newblock Chain-of-thought prompting elicits reasoning in large language models.
\newblock \emph{NeurIPS}, 2022.

\bibitem[Wu et~al.(2023{\natexlab{a}})Wu, Yin, Qi, Wang, Tang, and Duan]{wu2023visual}
Chenfei Wu, Shengming Yin, Weizhen Qi, Xiaodong Wang, Zecheng Tang, and Nan Duan.
\newblock Visual chatgpt: Talking, drawing and editing with visual foundation models.
\newblock \emph{arXiv preprint arXiv:2303.04671}, 2023{\natexlab{a}}.

\bibitem[Wu et~al.(2023{\natexlab{b}})Wu, Bansal, Zhang, Wu, Zhang, Zhu, Li, Jiang, Zhang, and Wang]{wu2023autogen}
Qingyun Wu, Gagan Bansal, Jieyu Zhang, Yiran Wu, Shaokun Zhang, Erkang Zhu, Beibin Li, Li~Jiang, Xiaoyun Zhang, and Chi Wang.
\newblock Autogen: Enabling next-gen llm applications via multi-agent conversation framework, 2023{\natexlab{b}}.

\bibitem[Wu et~al.(2023{\natexlab{c}})Wu, Jia, Zhang, Wu, Li, Zhu, Wang, Lee, Peng, and Wang]{wu2023empirical}
Yiran Wu, Feiran Jia, Shaokun Zhang, Qingyun Wu, Hangyu Li, Erkang Zhu, Yue Wang, Yin~Tat Lee, Richard Peng, and Chi Wang.
\newblock An empirical study on challenging math problem solving with gpt-4, 2023{\natexlab{c}}.

\bibitem[Yao et~al.(2022)Yao, Zhao, Yu, Du, Shafran, Narasimhan, and Cao]{yao2022react}
Shunyu Yao, Jeffrey Zhao, Dian Yu, Nan Du, Izhak Shafran, Karthik Narasimhan, and Yuan Cao.
\newblock React: Synergizing reasoning and acting in language models, 2022.

\bibitem[Yao et~al.(2024)Yao, Yu, Zhao, Shafran, Griffiths, Cao, and Narasimhan]{yao2024tree}
Shunyu Yao, Dian Yu, Jeffrey Zhao, Izhak Shafran, Tom Griffiths, Yuan Cao, and Karthik Narasimhan.
\newblock Tree of thoughts: Deliberate problem solving with large language models.
\newblock \emph{NeurIPS}, 2024.

\bibitem[Yuan et~al.(2023)Yuan, Chen, Wang, Fung, Peng, and Ji]{yuan2023craft}
Lifan Yuan, Yangyi Chen, Xingyao Wang, Yi~R. Fung, Hao Peng, and Heng Ji.
\newblock Craft: Customizing llms by creating and retrieving from specialized toolsets, 2023.

\bibitem[Yuan et~al.(2024)Yuan, Song, Chen, Tan, Shen, Kan, Li, and Yang]{yuan2024easytool}
Siyu Yuan, Kaitao Song, Jiangjie Chen, Xu~Tan, Yongliang Shen, Ren Kan, Dongsheng Li, and Deqing Yang.
\newblock Easytool: Enhancing llm-based agents with concise tool instruction, 2024.

\bibitem[Zhang et~al.(2024{\natexlab{a}})Zhang, Yang, Hu, Wang, Li, Sun, Zhang, Zhang, Liu, Zhu, Chang, Zhang, Yin, Liang, and Yang]{zhang2024proagent}
Ceyao Zhang, Kaijie Yang, Siyi Hu, Zihao Wang, Guanghe Li, Yihang Sun, Cheng Zhang, Zhaowei Zhang, Anji Liu, Song-Chun Zhu, Xiaojun Chang, Junge Zhang, Feng Yin, Yitao Liang, and Yaodong Yang.
\newblock Pro{A}gent: Building proactive cooperative agents with large language models.
\newblock In \emph{AAAI}, 2024{\natexlab{a}}.

\bibitem[Zhang et~al.(2024{\natexlab{b}})Zhang, Zhang, Ren, Li, and Yang]{zhang2024mlcopilot}
Lei Zhang, Yuge Zhang, Kan Ren, Dongsheng Li, and Yuqing Yang.
\newblock Mlcopilot: Unleashing the power of large language models in solving machine learning tasks, 2024{\natexlab{b}}.

\bibitem[Zhang et~al.(2023)Zhang, Shen, Lu, and Zhuang]{zhang2023data}
Wenqi Zhang, Yongliang Shen, Weiming Lu, and Yueting Zhuang.
\newblock Data-copilot: Bridging billions of data and humans with autonomous workflow.
\newblock \emph{arXiv preprint arXiv:2306.07209}, 2023.

\bibitem[Zhang et~al.(2024{\natexlab{c}})Zhang, Jiang, Han, Chen, Yang, and Ren]{zhang2024benchmarking}
Yuge Zhang, Qiyang Jiang, Xingyu Han, Nan Chen, Yuqing Yang, and Kan Ren.
\newblock Benchmarking data science agents, 2024{\natexlab{c}}.

\bibitem[Zheng et~al.(2021)Zheng, Zhang, Hong, Li, Tang, Xiong, Zhou, Wang, Sun, Zhu, et~al.]{zheng2021evolving}
Xiawu Zheng, Yang Zhang, Sirui Hong, Huixia Li, Lang Tang, Youcheng Xiong, Jin Zhou, Yan Wang, Xiaoshuai Sun, Pengfei Zhu, et~al.
\newblock Evolving fully automated machine learning via life-long knowledge anchors.
\newblock \emph{TPAMI}, 2021.

\bibitem[Zhou et~al.(2023)Zhou, Wang, Lu, Shi, Luo, Qin, Lu, Jia, Song, Zhan, et~al.]{zhou2023solving}
Aojun Zhou, Ke~Wang, Zimu Lu, Weikang Shi, Sichun Luo, Zipeng Qin, Shaoqing Lu, Anya Jia, Linqi Song, Mingjie Zhan, et~al.
\newblock Solving challenging math word problems using gpt-4 code interpreter with code-based self-verification, 2023.

\bibitem[Zhuge et~al.(2023)Zhuge, Liu, Faccio, Ashley, Csord{\'a}s, Gopalakrishnan, Hamdi, Hammoud, Herrmann, Irie, et~al.]{zhuge2023mindstorms}
Mingchen Zhuge, Haozhe Liu, Francesco Faccio, Dylan~R Ashley, R{\'o}bert Csord{\'a}s, Anand Gopalakrishnan, Abdullah Hamdi, Hasan Abed Al~Kader Hammoud, Vincent Herrmann, Kazuki Irie, et~al.
\newblock Mindstorms in natural language-based societies of mind, 2023.

\bibitem[Zhuge et~al.(2024)Zhuge, Wang, Kirsch, Faccio, Khizbullin, and Schmidhuber]{zhuge2024language}
Mingchen Zhuge, Wenyi Wang, Louis Kirsch, Francesco Faccio, Dmitrii Khizbullin, and Jurgen Schmidhuber.
\newblock Language agents as optimizable graphs, 2024.

\end{thebibliography}
\bibliographystyle{iclr2025_conference}

\appendix
\clearpage
\section{Limitations}\label{limitations}
\textbf{Diversity and complexity insufficient.}
Our novel framework \model{} outperforms other open-source frameworks on machine learning problems, 
yet are limited to entry-level Kaggle datasets and benchmarked against the capabilities of a junior human data scientist. These datasets are relatively small (under 500MB), with a limited number of columns (in the hundreds) and rows (in the tens of thousands), and mainly involve classification and regression tasks (as described in~\Cref{appendix: ml-benchmark-dsp}). However, we have not yet evaluated our \model{} on more challenging datasets involving large-scale data or complex tasks such as time series analysis, multi-label classification, or multi-table problems. In our future work, we plan to expand our dataset collection to include these types of problems to thoroughly evaluate our framework's performance and capabilities.
\textbf{Precise self-improvement}
Human data scientists usually perform multiple experiments on a dataset, focusing on pipeline optimization and hyperparameter tuning~\cite{liu2021winning, hutter2019automated}. 
Our \model{} integrates experience to enhance the node generation quality. The experience primarily involves tracking the progress of tasks and code.
However, it does not use numerical feedback from multiple experiences to develop and refine specific strategies, such as increasing the learning rate or using an ensemble technique, to improve the performance continuously for a given dataset, thus lacking the capability for automatic self-improvement. In the future, we aim to address this limitation by developing mechanisms that allow our model to conduct multiple experiments and derive insights from the numerical feedback for a given dataset on its own.
\textbf{Full-scale evaluation on mathematical problems.}
For the MATH problem, our experiments are limited to level-5 problems, primarily due to the budget constraints, we will explore more cost-effective strategies for evaluating our \model{} on a wider range of mathematical problems in future studies.
\section{Broader impact}\label{broader impacts}
Our work has the potential to significantly reduce the costs associated with a wide range of customized data science tasks, empowering professionals in the field to enhance their automation capabilities and efficiency. However, the flexibility of tools integration, while convenient for local code snippets integration, comes with potential risks. For instance, if users provide malicious code intended for unauthorized system penetration or web attacks, it could lead to security vulnerabilities. In our experiments, we mitigate this risk by prompting our \model{} to check the codes before generating new codes. Additional saftguards against these risks include collaborating exclusively with LLMs that adhere to robust safety policies. 

\clearpage

\section{Implemtation details}
\subsection{Programmable Node Generation}\label{appendix:tool_selection}
We illustrate the process of node generation process with tools.

\begin{figure*}[ht]
    \centering
    \includegraphics[width=\textwidth]{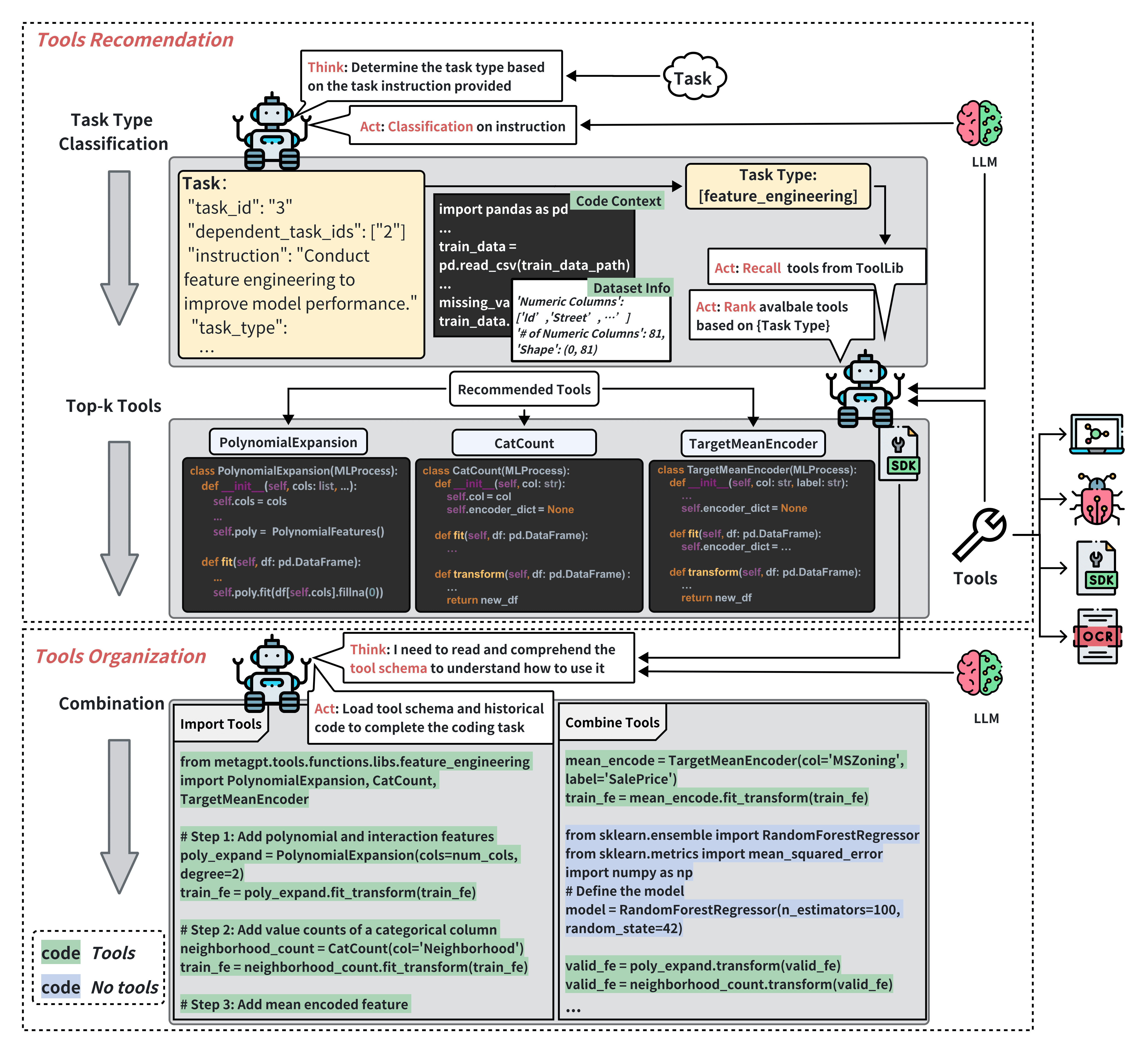}
    \caption{\textbf{Node generation pipeline in \model{}.} Tools are initially selected based on task metadata classification, followed by tools organization process which combines multiple tools as necessary to accomplish the tasks.}
    \label{fig:tool-usage}
\end{figure*}

\clearpage

\subsubsection{An example of tool schema}\label{sec:tool_schema}
Below is an example of tool schema we design in our framework.
\begin{tcolorbox}[title={\textbf{\small Tool schema for a feature engineering tool}}, colback=whitesmoke, colframe=darkblue, boxrule=2pt, arc=0mm]
{\scriptsize
\lstset{
  breaklines=true,  
  basicstyle=\ttfamily,  
}
\begin{lstlisting}
type: class
description: Add value counts of a categorical column as new feature.
methods:
  __init__:
    type: function
    description: Initialize self.
    parameters:
      properties:
        col:
          type: str
          description: Column for value counts.
      required:
      - col
  fit:
    type: function
    description: Fit a model to be used in subsequent transform.
    parameters:
      properties:
        df:
          type: pd.DataFrame
          description: The input DataFrame.
      required:
      - df
  fit_transform:
    type: function
    description: Fit and transform the input DataFrame.
    parameters:
      properties:
        df:
          type: pd.DataFrame
          description: The input DataFrame.
      required:
      - df
    returns:
    - type: pd.DataFrame
      description: The transformed DataFrame.
  transform:
    type: function
    description: Transform the input DataFrame with the fitted model.
    parameters:
      properties:
        df:
          type: pd.DataFrame
          description: The input DataFrame.
      required:
      - df
    returns:
    - type: pd.DataFrame
      description: The transformed DataFrame.

\end{lstlisting}
}
\end{tcolorbox}

\subsubsection{Tools details}
The tools of our Data Interpreter are listed in  \Cref{tab:tools}
\begin{table*}[t]
\tablestyle{1.8pt}{1.05}
\caption{
\textbf{Tools of our Data Interpreter.}}
\renewcommand\tabcolsep{20pt}
\renewcommand\arraystretch{1.2}
\small
\centering
\begin{tabular}{l|ccc}
\hline

\hline

\hline

\hline
\multicolumn{1}{l|}{Tool name} & \multicolumn{1}{c}{Tool type} & \multicolumn{1}{c}{Functions} & \multicolumn{1}{c}{Domain} \\
\midrule
\multicolumn{1}{l|}{FillMissingValue} & \multicolumn{1}{c}{Class} & \multicolumn{1}{c}{4} & \multicolumn{1}{c}{Machine learning} \\
\multicolumn{1}{l|}{MinMaxScale} & \multicolumn{1}{c}{Class} & \multicolumn{1}{c}{4} & \multicolumn{1}{c}{Machine learning} \\
\multicolumn{1}{l|}{StandardScale} & \multicolumn{1}{c}{Class} & \multicolumn{1}{c}{4} & \multicolumn{1}{c}{Machine learning} \\
\multicolumn{1}{l|}{MaxAbsScale} & \multicolumn{1}{c}{Class} & \multicolumn{1}{c}{4} & \multicolumn{1}{c}{Machine learning} \\
\multicolumn{1}{l|}{LabelEncode} & \multicolumn{1}{c}{Class} & \multicolumn{1}{c}{4} & \multicolumn{1}{c}{Machine learning} \\
\multicolumn{1}{l|}{OneHotEncode} & \multicolumn{1}{c}{Class} & \multicolumn{1}{c}{4} & \multicolumn{1}{c}{Machine learning} \\
\multicolumn{1}{l|}{OrdinalEncode} & \multicolumn{1}{c}{Class} & \multicolumn{1}{c}{4} & \multicolumn{1}{c}{Machine learning} \\
\multicolumn{1}{l|}{RobustScale} & \multicolumn{1}{c}{Class} & \multicolumn{1}{c}{4} & \multicolumn{1}{c}{Machine learning} \\
\multicolumn{1}{l|}{CatCount} & \multicolumn{1}{c}{Class} & \multicolumn{1}{c}{4} & \multicolumn{1}{c}{Machine learning} \\
\multicolumn{1}{l|}{TargetMeanEncoder} & \multicolumn{1}{c}{Class} & \multicolumn{1}{c}{4} & \multicolumn{1}{c}{Machine learning} \\
\multicolumn{1}{l|}{KFoldTargetMeanEncoder} & \multicolumn{1}{c}{Class} & \multicolumn{1}{c}{4} & \multicolumn{1}{c}{Machine learning} \\
\multicolumn{1}{l|}{CatCross} & \multicolumn{1}{c}{Class} & \multicolumn{1}{c}{5} & \multicolumn{1}{c}{Machine learning} \\
\multicolumn{1}{l|}{SplitBins} & \multicolumn{1}{c}{Class} & \multicolumn{1}{c}{4} & \multicolumn{1}{c}{Machine learning} \\
\multicolumn{1}{l|}{GeneralSelection} & \multicolumn{1}{c}{Class} & \multicolumn{1}{c}{4} & \multicolumn{1}{c}{Machine learning} \\
\multicolumn{1}{l|}{TreeBasedSelection} & \multicolumn{1}{c}{Class} & \multicolumn{1}{c}{4} & \multicolumn{1}{c}{Machine learning} \\
\multicolumn{1}{l|}{VarianceBasedSelection} & \multicolumn{1}{c}{Class} & \multicolumn{1}{c}{4} & \multicolumn{1}{c}{Machine learning} \\
\multicolumn{1}{l|}{PolynomialExpansion} & \multicolumn{1}{c}{Class} & \multicolumn{1}{c}{4} & \multicolumn{1}{c}{Machine learning} \\

\midrule

\multicolumn{1}{l|}{GPTvGenerator} & \multicolumn{1}{c}{Class} & \multicolumn{1}{c}{3} & \multicolumn{1}{c}{Multimodal} \\
\multicolumn{1}{l|}{SDEngine} & \multicolumn{1}{c}{Class} & \multicolumn{1}{c}{5} & \multicolumn{1}{c}{Multimodal} \\

\midrule

\multicolumn{1}{l|}{scrape\_web\_playwright} & \multicolumn{1}{c}{Function} & \multicolumn{1}{c}{1} & \multicolumn{1}{c}{Common} \\
\hline

\hline

\hline

\hline
\end{tabular}
    \label{tab:tools}
\end{table*}

\subsubsection{Tool usage prompts}
We use two types of prompts for tool utilization. For open-ended tasks, we use zero-shot prompts, and for machine-learning tasks, we use one-shot prompts as illustrated below.
\begin{tcolorbox}[title={\textbf{\small  Zero-shot tool usage prompt}}, 
colback=whitesmoke, colframe=darkblue, boxrule=2pt, arc=0mm]
{\scriptsize
\lstset{
  breaklines=true,  
  basicstyle=\ttfamily,  
}
\begin{lstlisting}
# Instruction
Write complete code for 'Current Task'. And avoid duplicating code from finished tasks, such as repeated import of packages, reading data, etc.
Specifically, {tool_type_usage_prompt}

# Capabilities
- You can utilize pre-defined tools in any code lines from 'Available Tools' in the form of Python Class.
- You can freely combine the use of any other public packages, like sklearn, numpy, pandas, etc..

# Available Tools (can be empty):
Each Class tool is described in JSON format. When you call a tool, import the tool first.
{tool_schemas}

# Constraints:
- Ensure the output new code is executable in the same Jupyter notebook with the previous tasks code has been executed.
- Always prioritize using pre-defined tools for the same functionality.
\end{lstlisting}
}
\end{tcolorbox}

\clearpage
\begin{tcolorbox}[title={\textbf{\small One-shot tool usage prompt}}, 
colback=whitesmoke, colframe=darkblue, boxrule=2pt, arc=0mm, breakable]
{\scriptsize
\lstset{
  breaklines=true,  
  basicstyle=\ttfamily,  
}
\begin{lstlisting}
# Capabilities
- You can utilize pre-defined tools in any code lines from 'Available Tools' in the form of Python Class.
- You can freely combine the use of any other public packages, like sklearn, numpy, pandas, etc..

# Available Tools:
Each Class tool is described in JSON format. When you call a tool, import the tool from its path first.
{tool_schemas}

# Output Example:
when the current task is "do data preprocess, like fill missing value, handle outliers, etc.", the code can be like:
```python
# Step 1: fill missing value
# Tools used: ['FillMissingValue']
from metagpt.tools.libs.data_preprocess import FillMissingValue

train_processed = train.copy()
test_processed = test.copy()
num_cols = train_processed.select_dtypes(include='number').columns.tolist()
if 'label' in num_cols:
    num_cols.remove('label')
fill_missing_value = FillMissingValue(features=num_cols, strategy='mean')
fill_missing_value.fit(train_processed)
train_processed = fill_missing_value.transform(train_processed)
test_processed = fill_missing_value.transform(test_processed)

# Step 2: handle outliers
for col in num_cols:
    low, high = train_processed[col].quantile([0.01, 0.99])
    train_processed[col] = train_processed[col].clip(low, high)
    test_processed[col] = test_processed[col].clip(low, high)
```end

# Constraints:
- Ensure the output new code is executable in the same Jupyter notebook with the previous tasks code has been executed.
- Always prioritize using pre-defined tools for the same functionality.
- Always copy the DataFrame before processing it and use the copy to process.
\end{lstlisting}
}
\end{tcolorbox}

\section{Experiment details}
\subsection{Dataset} \label{appendix:exp-datasets}
\paragraph{InfiAgent-DABench} \label{appendix:infiagent}
InfiAgent-DABench focuses on evaluating the data analysis capabilities of agents. It comprises 257 data analysis problems, categorized into the following seven areas and their combinations: summary statistics, feature engineering, correlation analysis, machine learning, distribution analysis, outlier detection, and comprehensive data preprocessing. Each category includes problems of varying difficulty levels. Below, we present some specific prompt cases to provide an intuitive understanding of the task settings in InfiAgent-DABench.

\begin{tcolorbox}[title={\textbf{\small InfiAgent-DABench prompt}}, 
colback=whitesmoke, colframe=darkblue, boxrule=2pt, arc=0mm, breakable]
{\scriptsize
\lstset{
  breaklines=true,  
  basicstyle=\ttfamily,  
}
\begin{lstlisting}
1. category: ['Summary Statistics'] , level: easy. 
 prompt: Please write a Python code snippet to Calculate the mean and standard deviation of the abs_diffsel column. based on the following details: The task is to { The mean and standard deviation should be calculated directly from the 'abs_diffsel' column. Do not remove any outliers or modify the data prior to calculation. The mean and standard deviation should be computed directly from all available data points. }  and formatted as { @mean[mean_value] @std_dev[std_dev_value] where "mean_value" is a positive float number, rounded to two decimal places. where "std_dev_value" is a positive float number, rounded to two decimal places.. The data is stored in a file saved in "InfiAgent/examples/DA-Agent/data/da-dev-tables/ferret-Pitt-2-preinf-lib2-100_sitediffsel.csv", and the difficulty level is easy. 

 
2. category: ['Feature Engineering', 'Correlation Analysis'] , level: medium. 
prompt: Please write a Python code snippet to Create a new feature called 'FamilySize' by combining the 'SibSp' and 'Parch' columns, which represents the total number of family members a passenger had aboard the Titanic. Then, find the correlation coefficient between 'FamilySize' and 'Survived'. based on the following details: The task is to Create 'FamilySize' by adding up 'SibSp' and 'Parch', then calculate the Pearson correlation coefficient between 'FamilySize' and 'Survived'.  and formatted as @correlation_coefficient[number] where "number" is the calculated Pearson correlation coefficient between 'FamilySize' and 'Survived', rounded to two decimal places.. The data is stored in a file saved in "InfiAgent/examples/DA-Agent/data/da-dev-tables/titanic.csv", and the difficulty level is medium.  

3. category: ['Comprehensive Data Preprocessing', 'Distribution Analysis'] , level: hard. 
 prompt: Please write a Python code snippet to 2. Preprocess the dataset by handling missing values in the "24-Hour Passes Purchased (midnight to 11:59 pm)" and "7-Day Passes Purchased (midnight to 11:59 pm)" columns. Use the mean imputation method to fill in the missing values. Then, analyze the distribution of the "Trips over the past 24-hours (midnight to 11:59pm)" column before and after the missing value imputation process. Evaluate if the imputation has significantly affected the distribution and what implications it has on the dataset analysis. based on the following details: The task is to Use the mean imputation method to fill in missing values for both the "24-Hour Passes Purchased (midnight to 11:59 pm)" and "7-Day Passes Purchased (midnight to 11:59 pm)" columns. Then, calculate the mean, median, standard deviation, skewness, and kurtosis for the "Trips over the past 24-hours (midnight to 11:59pm)" column before and after imputation.  and formatted as @pre_mean[mean_before] @pre_median[median_before] @pre_sd[sd_before] @pre_skewness[skew_before] @pre_kurtosis[kurt_before] @post_mean[mean_after] @post_median[median_after] @post_sd[sd_after] @post_skewness[skew_after] @post_kurtosis[kurt_after] where all variables represent the corresponding statistical values calculated before (prefix: pre) and after (prefix: post) the imputation, each rounded to two decimal places.. The data is stored in a file saved in "InfiAgent/examples/DA-Agent/data/da-dev-tables/2014_q4.csv", and the difficulty level is hard. 
 



\end{lstlisting}
}
\end{tcolorbox}

\paragraph{ML-Benchmark}
This dataset encompassed eight representative machine learning tasks categorized into three difficulty levels, ranging from easy (level 1) to most complex (level 3). 
Each task was accompanied by data, a concise description, standard user requirements, suggested steps, and metrics (see \Cref{tab:dataset-info} in the Appendix). 
For tasks labeled as ``toy", the data was not divided into training and test splits, which required the framework to perform data splitting during modeling.
\paragraph{Open-ended task benchmark}
To evaluate the ability to generalize to real-world tasks, we developed the Open-ended task benchmark, comprising 20 tasks.
Each task required the framework to understand user needs, break down complex tasks, and execute code. 
They delineated their requirements, foundational data or sources, steps for completion, and specific metrics. 
The scope was broad, encompassing common needs like Optical Character Recognition (OCR), web search and crawling (WSC), automated email replies (ER), web page imitation (WPI), text-to-image conversion (T2I), image-to-HTML code generation (I2C), image background removal (IBR), and mini-game generation (MGG). We showcase about these tasks in~\Cref{fig:open_ended_task_details_task_1_7},~\Cref{fig:open_ended_task_details_task_14_17}, and~\Cref{fig:open_ended_task_details_task_18_20} in the Appendix.

\paragraph{MATH dataset}
The MATH dataset~\cite{hendrycks2021measuring} comprises 12,500 problems, with 5,000 designated as the test set, covering various subjects and difficulty levels. 
These subjects include Prealgebra (Prealg), Algebra, Number Theory (N.Theory), Counting and Probability (C.Prob), Geometry, Intermediate Algebra, and Precalculus (Precalc), with problems categorized from levels "1" to "5" based on difficulty. 
Following the setting of Wu et al.~\cite{wu2023empirical}, we evaluated four typical problem types (C.Prob, N.Theory, Prealg, Precalc), excluding level-5 geometry problems from the test set.

\subsection{Evaluation metrics} \label{appendix:exp-metrics}
In the MATH benchmark~\cite{hendrycks2021measuring}, accuracy served as the chosen evaluation metric, aligning with the setting proposed in \cite{wu2023empirical,hendrycks2021measuring}. 


For the ML-Benchmark, three evaluation metrics were utilized: completion rate (CR), normalized performance score (NPS), and comprehensive score (CS). These metrics provided comprehensive insights into the model's performance and were defined as follows:

\textit{\textbf{Completion rate (CR)}}: 
In the task requirements description, there were $T$ steps, and the task completion status of each step was denoted by a score $s_t$, with a maximum score $s_{max}$ of 2 and a minimum score $s_{min}$ of 0. 
The task completion status categories were defined as follows: missing (score of 0), fail (score of 0), success - non-compliant (score of 1), success-compliant (score of 2), and optional step (not involved in scoring). 
To measure the completion level, we proposed a completion ratio where the numerator was the sum of scores $s_t$ for each step, and the denominator was the sum of the maximum possible scores for all steps ($s_{max} \times T$):
\begin{align}
\text{CR} &= \frac{\sum_{t=1}^{T} s_t}{s_{max} \times T}.
\label{equ:CompletionRate}
\end{align}

\textit{\textbf{Normalized performance score (NPS)}}: 
In our ML-Benchmark, each task was associated with its evaluation metric, which may vary between tasks, including metrics such as accuracy, F1, AUC and RMSLE, etc. For metrics such as accuracy, F1, and AUC, we presented the raw values to facilitate comparison across identical data tasks.
We normalize all performance values $s$:
\begin{align}
\text{NPS} = 
\begin{cases} 
\dfrac{1}{1 + s}, & \text{if } s \text{ is smaller the better} \\
s, & \text{otherwise}.
\end{cases}
\label{equ:NormalizedMetric}
\end{align}
This transformation ensured that loss-based metrics like RMSLE are scaled from 0 to 1, with higher normalized performance score values indicating better performance.

\textit{\textbf{Comprehensive score (CS)}}: To simultaneously assess both the completion rate of task requirements and the performance of generated machine learning models, we calculated the weighted sum of CR and NPS as follows:
\begin{align}
    \text{CS} &= 0.5 \times \text{CR} + 0.5 \times \text{NPS}.
    \label{equ:ComprehensiveScore}
\end{align}
Considering the lack of unified performance standards for open-ended tasks, we default to $\text{NPS}=0$ and directly equate CS to CR.


\subsection{Additional results}
\subsubsection{Additional results of ML-benchmark and Math dataset}
For a deeper understanding, \Cref{tab:full-ml-score} presents the results on the ML-benchmark for both Completion Rate and Normalized Performance Score metrics. Additionally, \Cref{tab:full-ml-score-abl} showcases the results of ablation experiments on the ML-benchmark, focusing on the completion rate (CR) and normalized performance score (NPS).

\begin{table*}[ht]
\tablestyle{1.8pt}{1.05}
\caption{
\textbf{Additional performance comparisons on ML benchmark.} ``WR", ``BCW", ``ICR", ``SCTP", and ``SVPC"  represent ``Wine recognition"", ``Breast cancer wisconsin", ``ICR - Identifying age-related conditions", ``Santander customer transaction prediction", and ``Santander value prediction challenge", respectively. ``Avg." denotes ``Average".
}

\renewcommand\tabcolsep{6.0pt}
\renewcommand\arraystretch{1.2}
\small
\centering
\begin{tabular}{l|cccccccccc}
\hline

\hline

\hline

\hline
\multicolumn{1}{l|}{Model / Task} & \multicolumn{1}{c}{WR} & \multicolumn{1}{c}{BCW} & \multicolumn{1}{c}{Titanic}
& \multicolumn{1}{c}{House Prices} & \multicolumn{1}{c}{SCTP} & \multicolumn{1}{c}{ICR} & \multicolumn{1}{c}{SVPC} & \multicolumn{1}{|c}{Avg.}
\\
\midrule
\multicolumn{3}{l}{\textit{Completion rate}} \\ 
\midrule

\multicolumn{1}{l|}{AutoGen} & \multicolumn{1}{c}{0.92} & \multicolumn{1}{c}{\textbf{1.00}} & \multicolumn{1}{c}{0.92}
& \multicolumn{1}{c}{0.83} & \multicolumn{1}{c}{0.83} & \multicolumn{1}{c}{0.83} & \multicolumn{1}{c}{0.83} & \multicolumn{1}{|c}{0.88}
\\

\multicolumn{1}{l|}{OpenInterpreter} & \multicolumn{1}{c}{\textbf{1.00}} & \multicolumn{1}{c}{0.90}
 & \multicolumn{1}{c}{0.92} & \multicolumn{1}{c}{0.88} & \multicolumn{1}{c}{0.85} & \multicolumn{1}{c}{0.91} & \multicolumn{1}{c}{0.88} & \multicolumn{1}{|c}{0.90}
\\

\multicolumn{1}{l|}{TaskWeaver} & \multicolumn{1}{c}{\textbf{1.00}} & \multicolumn{1}{c}{\textbf{1.00}} & \multicolumn{1}{c}{0.83}
& \multicolumn{1}{c}{0.88} & \multicolumn{1}{c}{0.67} & \multicolumn{1}{c}{0.83} & \multicolumn{1}{c}{0.80} & \multicolumn{1}{|c}{0.86}
\\

\multicolumn{1}{l|}{XAgent} &
\multicolumn{1}{c}{\textbf{1.00}} & \multicolumn{1}{c}{\textbf{1.00}} & \multicolumn{1}{c}{0.83}
& \multicolumn{1}{c}{0.83} & \multicolumn{1}{c}{0} & \multicolumn{1}{c}{0.67} & \multicolumn{1}{c}{0} & \multicolumn{1}{|c}{0.62}
\\

\multicolumn{1}{l|}{OpenDevin} &
\multicolumn{1}{c}{\textbf{1.00}} & \multicolumn{1}{c}{\textbf{1.00}} & \multicolumn{1}{c}{0.92}
& \multicolumn{1}{c}{1.00} & \multicolumn{1}{c}{1.00} & \multicolumn{1}{c}{0.83} & \multicolumn{1}{c}{1.00} & \multicolumn{1}{|c}{0.96}
\\

\rowcolor[gray]{.9}
\multicolumn{1}{l|}{\textbf{Data Interpreter}} &
\multicolumn{1}{c}{\textbf{1.00}} & \multicolumn{1}{c}{\textbf{1.00}} & \multicolumn{1}{c}{\textbf{1.00}}
& \multicolumn{1}{c}{\textbf{1.00}} & \multicolumn{1}{c}{\textbf{1.00}} & \multicolumn{1}{c}{\textbf{1.00}} & \multicolumn{1}{c}{\textbf{1.00}} & \multicolumn{1}{|c}{\textbf{1.00}}
\\
\midrule
\multicolumn{3}{l}{\textit{Normalized performance score}} \\
\midrule
\multicolumn{1}{l|}{AutoGen}  & \multicolumn{1}{c}{\textbf{1.00}} & \multicolumn{1}{c}{0.97} & \multicolumn{1}{c}{0.82}
& \multicolumn{1}{c}{0.88} & \multicolumn{1}{c}{0.82} & \multicolumn{1}{c}{0.71} & \multicolumn{1}{c}{0.63} & \multicolumn{1}{|c}{0.83}
\\

\multicolumn{1}{l|}{OpenInterpreter} & \multicolumn{1}{c}{\textbf{1.00}} & \multicolumn{1}{c}{0.96} & \multicolumn{1}{c}{0.81} & \multicolumn{1}{c}{0.87} & \multicolumn{1}{c}{0.52} & \multicolumn{1}{c}{0.25}
& \multicolumn{1}{c}{0} & \multicolumn{1}{|c}{0.63} 
\\

\multicolumn{1}{l|}{TaskWeaver} &  \multicolumn{1}{c}{\textbf{1.00}} & \multicolumn{1}{c}{0.96} & \multicolumn{1}{c}{0.43}
& \multicolumn{1}{c}{0.49} & \multicolumn{1}{c}{0} & \multicolumn{1}{c}{0.65} & \multicolumn{1}{c}{0.17} & \multicolumn{1}{|c}{0.53}
\\
\multicolumn{1}{l|}{XAgent} & \multicolumn{1}{c}{\textbf{1.00}} & \multicolumn{1}{c}{0.94} & \multicolumn{1}{c}{0}
& \multicolumn{1}{c}{0} & \multicolumn{1}{c}{0} & \multicolumn{1}{c}{0} & \multicolumn{1}{c}{0} & \multicolumn{1}{|c}{0.28}
\\

\multicolumn{1}{l|}{OpenDevin} & \multicolumn{1}{c}{\textbf{0.96}} & \multicolumn{1}{c}{0.96} & \multicolumn{1}{c}{0.81}
& \multicolumn{1}{c}{0.87} & \multicolumn{1}{c}{0.86} & \multicolumn{1}{c}{0.62} & \multicolumn{1}{c}{0.45} & \multicolumn{1}{|c}{0.79}
\\

\rowcolor[gray]{.9}
\multicolumn{1}{l|}{\textbf{Data Interpreter}} & \multicolumn{1}{c}{0.96} & \multicolumn{1}{c}{\textbf{0.99}} & \multicolumn{1}{c}{\textbf{0.82}}
& \multicolumn{1}{c}{\textbf{0.91}} & \multicolumn{1}{c}{\textbf{0.89}} & \multicolumn{1}{c}{\textbf{0.91}} & \multicolumn{1}{c}{\textbf{0.77}} & \multicolumn{1}{|c}{\textbf{0.89}}
\\

\hline

\hline

\hline

\hline
\end{tabular}
    \label{tab:full-ml-score}
\end{table*}

\begin{table}[ht]

    \caption{
    \textbf{Additional performance comparisons on MATH dataset.} ``Avg." and ``Std." denotes ``Average", ``Standard Deviation" respectively.
    }
    \centering
    \renewcommand\tabcolsep{9.2pt}
    \renewcommand\arraystretch{1.2}
    \small
    \centering
    \begin{tabular}{l|cc|ccccc}
    \hline
    
    \hline
    
    \hline
    
    \hline
    
    \toprule
        \multirow{2}{*}{Category} & \multirow{2}{*}{MathChat} & \multirow{2}{*}{AutoGen} & \multicolumn{5}{c}{Data Interpreter} \\
        & & & Avg. & Trial1 & Trail2 & Trail3 & Std.(\%) \\
    \midrule
        C.Prob & 0.52 & 0.59 & 0.68 & 0.70 & 0.66 & 0.68 & 2.05 \\ 
        N.Theory & 0.60 & 0.66 & {0.82} & 0.81 & 0.82 & 0.82 & 0.99 \\
        Prealg & 0.60 & 0.63 & {0.74} & 0.73 & 0.75 & 0.75 & 1.20 \\
        Precalc & 0.19 & 0.12 & {0.29} & 0.28 & 0.30 & 0.29 & 1.13 \\ 
    \hline

\hline

\hline

\hline
    \end{tabular}
\end{table}

\subsubsection{Ablation study}
Here we provide detailed ablation study results on core modules. 

\begin{table*}[ht]
\tablestyle{1.5pt}{1.05}
\caption{
\textbf{Ablation on core modules.} Evaluated with CR, NPS and CS on ML-Benchmark. ``IGR" stands for Iterative Graph Refinement, and ``PNG" denotes Programmable Node Generation. ``ICR", ``SCTP", and ``SVPC" represent ``ICR - Identifying age-related conditions", ``Santander customer transaction prediction", and ``Santander value prediction challenge", respectively. }

\renewcommand\tabcolsep{5.9pt}
\renewcommand\arraystretch{1.05}

\small
\centering
\begin{tabular}{cccc|cccc}
\hline

\hline

\hline

\hline
\multicolumn{1}{c}{Code execution} & \multicolumn{1}{c}{IGR} &\multicolumn{1}{c|}{PNG} & \multicolumn{1}{c}{House Prices} & \multicolumn{1}{c}{SCTP} & \multicolumn{1}{c}{SVPC} &\multicolumn{1}{c}{ICR} & \multicolumn{1}{|c}{Avg.} \\

\midrule
\multicolumn{1}{c}{\textit{Completion rate}} \\ 
\midrule

\multicolumn{1}{c}{\checkmark} &\multicolumn{1}{c}{} & \multicolumn{1}{c}{} & \multicolumn{1}{c}{0.58} & \multicolumn{1}{c}{0.33} & \multicolumn{1}{c}{0.67} & \multicolumn{1}{c}{0.33} & \multicolumn{1}{|c}{0.48} \\
\multicolumn{1}{c}{\checkmark} &\multicolumn{1}{c}{\checkmark} & \multicolumn{1}{c|}{} & \multicolumn{1}{c}{1.00} & \multicolumn{1}{c}{1.00} & \multicolumn{1}{c}{0.92} & \multicolumn{1}{c}{0.88} & \multicolumn{1}{|c}{0.95} \\
\rowcolor[gray]{.9}
\multicolumn{1}{c}{\checkmark} &\multicolumn{1}{c}{\checkmark}  & \multicolumn{1}{c|}{\checkmark} & \multicolumn{1}{c}{\textbf{1.00}} & \multicolumn{1}{c}{\textbf{1.00}} & \multicolumn{1}{c}{\textbf{1.00}} & \multicolumn{1}{c}{\textbf{1.00}} & \multicolumn{1}{|c}{\textbf{1.00}} \\

\midrule
\multicolumn{1}{c}{\textit{Normalized performance score}} \\
\midrule
\multicolumn{1}{c}{\checkmark} &\multicolumn{1}{c}{} & \multicolumn{1}{c|}{} & \multicolumn{1}{c}{0.43} & \multicolumn{1}{c}{0} & \multicolumn{1}{c}{0.64} & \multicolumn{1}{c}{0} & \multicolumn{1}{|c}{0.27} \\
\multicolumn{1}{c}{\checkmark} &\multicolumn{1}{c}{\checkmark} & \multicolumn{1}{c|}{} & \multicolumn{1}{c}{0.91} & \multicolumn{1}{c}{0.82} & \multicolumn{1}{c}{0.68} & \multicolumn{1}{c}{0.60} & \multicolumn{1}{|c}{0.75} \\
\rowcolor[gray]{.9}
\multicolumn{1}{c}{\checkmark} &\multicolumn{1}{c}{\checkmark} & \multicolumn{1}{c|}{\checkmark} & \multicolumn{1}{c}{\textbf{0.91}} & \multicolumn{1}{c}{\textbf{0.89}} & \multicolumn{1}{c}{\textbf{0.77}} & \multicolumn{1}{c}{\textbf{0.91}} & \multicolumn{1}{|c}{\textbf{0.87}} \\

\midrule
\multicolumn{1}{c}{\textit{Comprehensive score}} \\
\midrule
\multicolumn{1}{c}{\checkmark} &\multicolumn{1}{c}{} & \multicolumn{1}{c|}{} & \multicolumn{1}{c}{0.51} & \multicolumn{1}{c}{0.17} & \multicolumn{1}{c}{0.66} & \multicolumn{1}{c}{0.17} & \multicolumn{1}{|c}{0.37} \\
\multicolumn{1}{c}{\checkmark} &\multicolumn{1}{c}{\checkmark} & \multicolumn{1}{c|}{} & \multicolumn{1}{c}{0.96} & \multicolumn{1}{c}{0.91} & \multicolumn{1}{c}{0.80} & \multicolumn{1}{c}{0.74} & \multicolumn{1}{|c}{0.85} \\

\rowcolor[gray]{.9}
\multicolumn{1}{c}{\checkmark} &\multicolumn{1}{c}{\checkmark} & \multicolumn{1}{c|}{\checkmark} & \multicolumn{1}{c}{\textbf{0.96}} & \multicolumn{1}{c}{\textbf{0.95}} & \multicolumn{1}{c}{\textbf{0.89}} & \multicolumn{1}{c}{\textbf{0.96}} & \multicolumn{1}{|c}{\textbf{0.94}} \\
\hline

\hline

\hline

\hline
\end{tabular}
\label{tab:full-ml-score-abl}
\end{table*}

\label{Sec:ab_study}

\clearpage
\section{Additional Examples}

\subsection{An Example of Task Graph}
\label{sec:plan_example}
Here is an example of a task graph. The user requirement is: ``This is a dataset featuring sensor readings from industrial machines, aimed at predicting machine operational status (normal or faulty). Visualize the analysis and prediction results with high-quality graphs. Train data path: \{train\_path\}, eval data path: \{eval\_path\}."
\begin{tcolorbox}[title={\textbf{\small Task graph example}}, 
colback=whitesmoke, colframe=darkblue, boxrule=2pt, arc=0mm, breakable]
{\scriptsize
\lstset{
  breaklines=true,  
  basicstyle=\ttfamily,  
}
\begin{lstlisting}
[ 
    { 
        "task_id": "1", 
        "dependent_task_ids": [], 
        "instruction": "Conduct data exploration on the train_sett.csv to understand its structure, missing values, and basic statistics." 
    }, 
    { 
        "task_id": "2", 
        "dependent_task_ids": ["1"], 
        "instruction": "Perform correlation analysis and causal inferences to identify relationships between variables." 
    }, 
    { 
        "task_id": "3", 
        "dependent_task_ids": ["1"], 
        "instruction": "Implement anomaly detection to identify and handle outliers in the dataset." 
    }, 
    { 
        "task_id": "4", 
        "dependent_task_ids": ["2", "3"], 
        "instruction": "Carry out feature engineering to prepare the dataset for predictive modeling." 
    }, 
    { 
        "task_id": "5", 
        "dependent_task_ids": ["4"], 
        "instruction": "Develop a predictive model using the processed dataset to determine machine operational status." 
    }, 
    { 
        "task_id": "6", 
        "dependent_task_ids": ["5"], 
        "instruction": "Evaluate the model's performance using the eval_set.csv" 
    }, 
    { 
        "task_id": "7", 
        "dependent_task_ids": ["5", "6"], 
        "instruction": "Visualize the analysis and prediction results with high-quality graphs." 
    }
]
\end{lstlisting}
}
\end{tcolorbox}

\subsection{Runtime results of task graph}
\begin{figure}[ht]
    \centering
        \includegraphics[width=\textwidth]
       {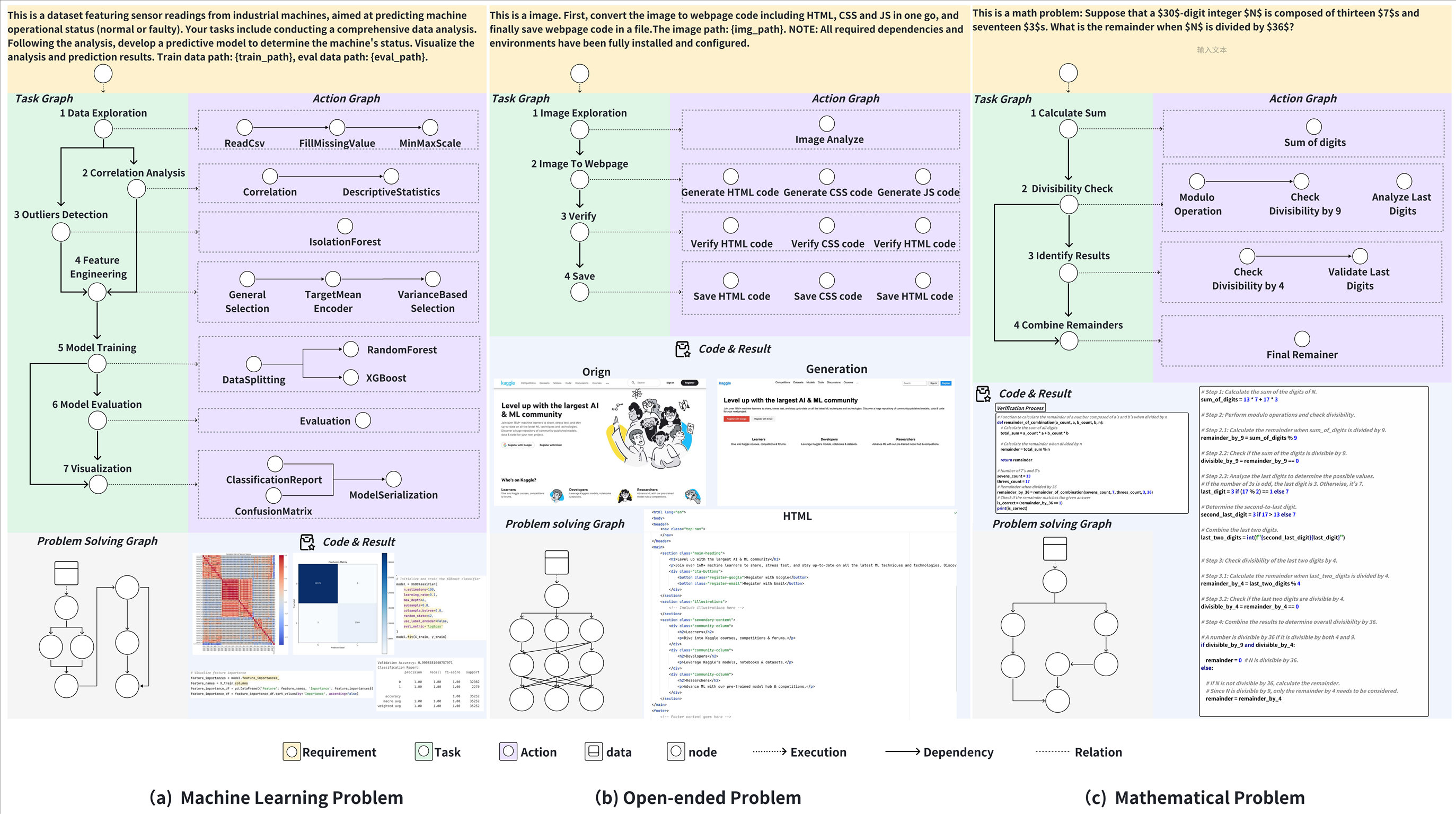}
            \caption{\textbf{Runtime examples of \model{}}: machine learning, webpage imitation, and math problem solving}
    \label{fig:graph_examples}
\end{figure}

We provide three distinct runtime results of our model, \model{}, to offer an in-depth demonstration of its capabilities. These results meticulously showcase the intricacies of the task graph, action graph, and the overall graph structure as shown in~\Cref{fig:graph_examples}.

\subsection{Additional results of Open-ended tasks}
\begin{figure}[th]
    \centering
    \includegraphics[width=\textwidth]{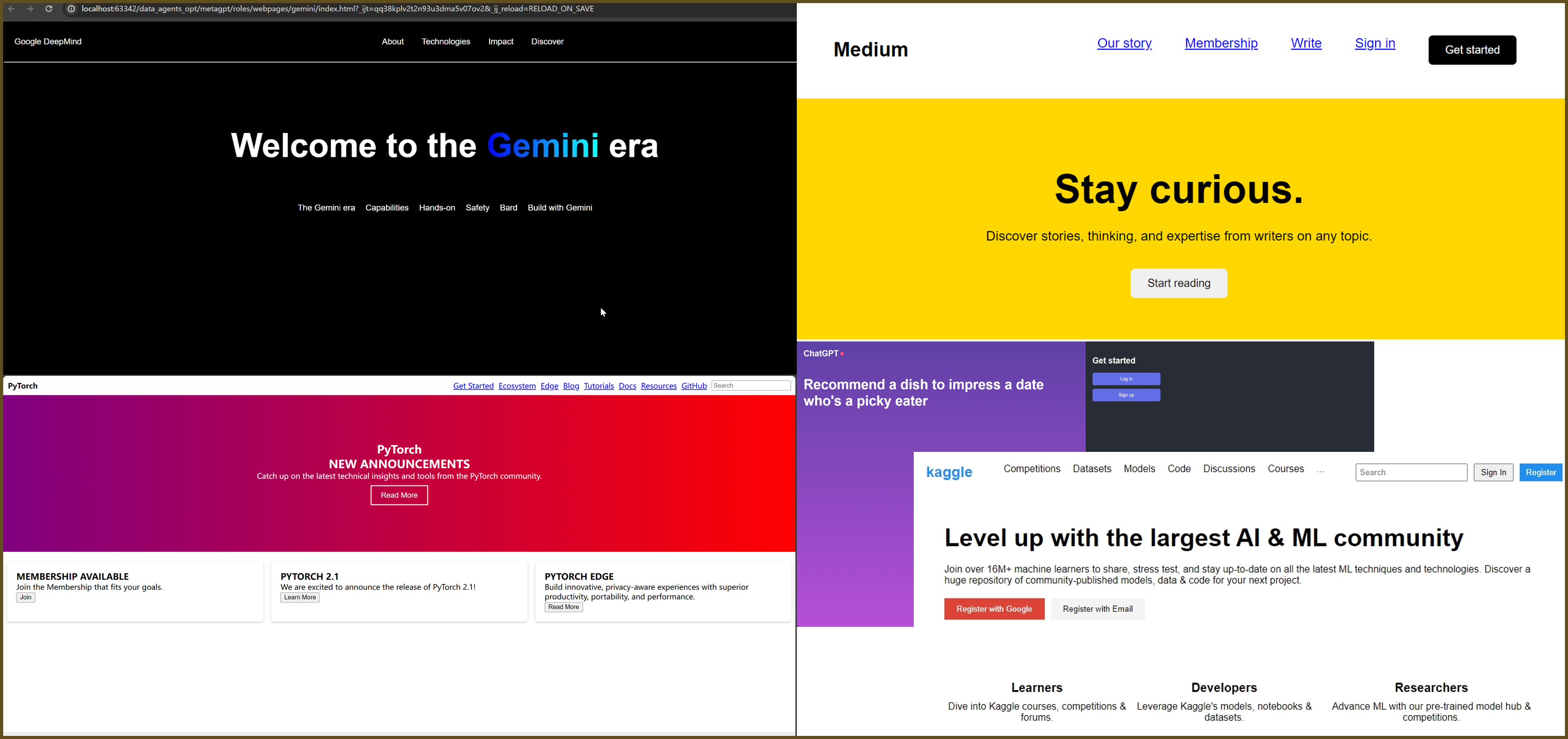}
    \caption{Web page imitation by \model{}}
    \label{fig:output-of-open-ended-tasks}
\end{figure}

\begin{figure}[th]
    \centering
    \includegraphics[width=\textwidth]{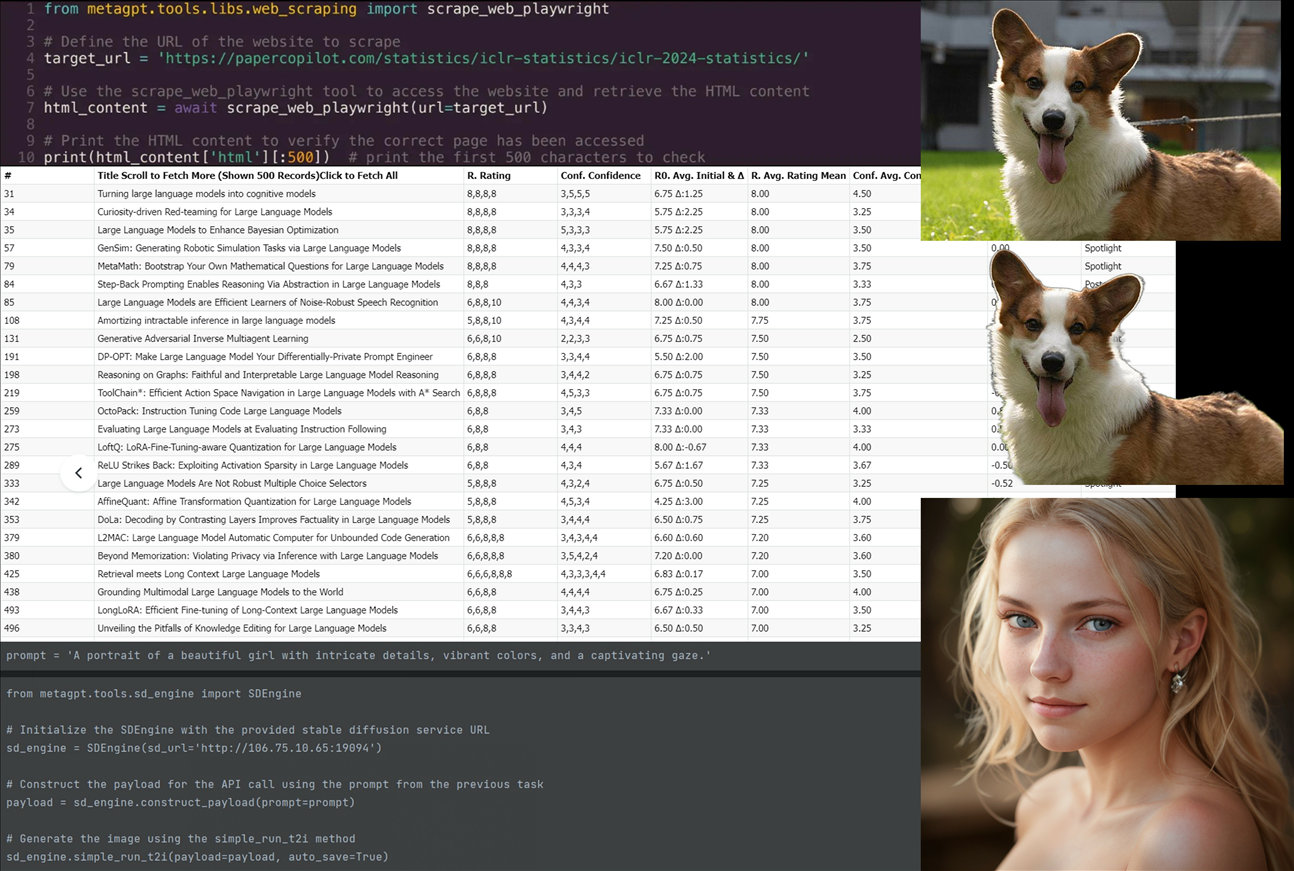}
    \caption{Image background removal / text-to-image / web search and crawling by \model{}}
    \label{fig:output-of-open-ended-tasks-multi-modal}
\end{figure}

We present the results by the \model{} of several open-ended tasks in two figures: tasks 8, 9, 10, and 13 in \Cref{fig:output-of-open-ended-tasks}, and tasks 4, 14, and 15 in \Cref{fig:output-of-open-ended-tasks-multi-modal}.
\subsection{Result of data visualization }
\begin{figure*}[th]
    \centering
    \includegraphics[width=\textwidth]{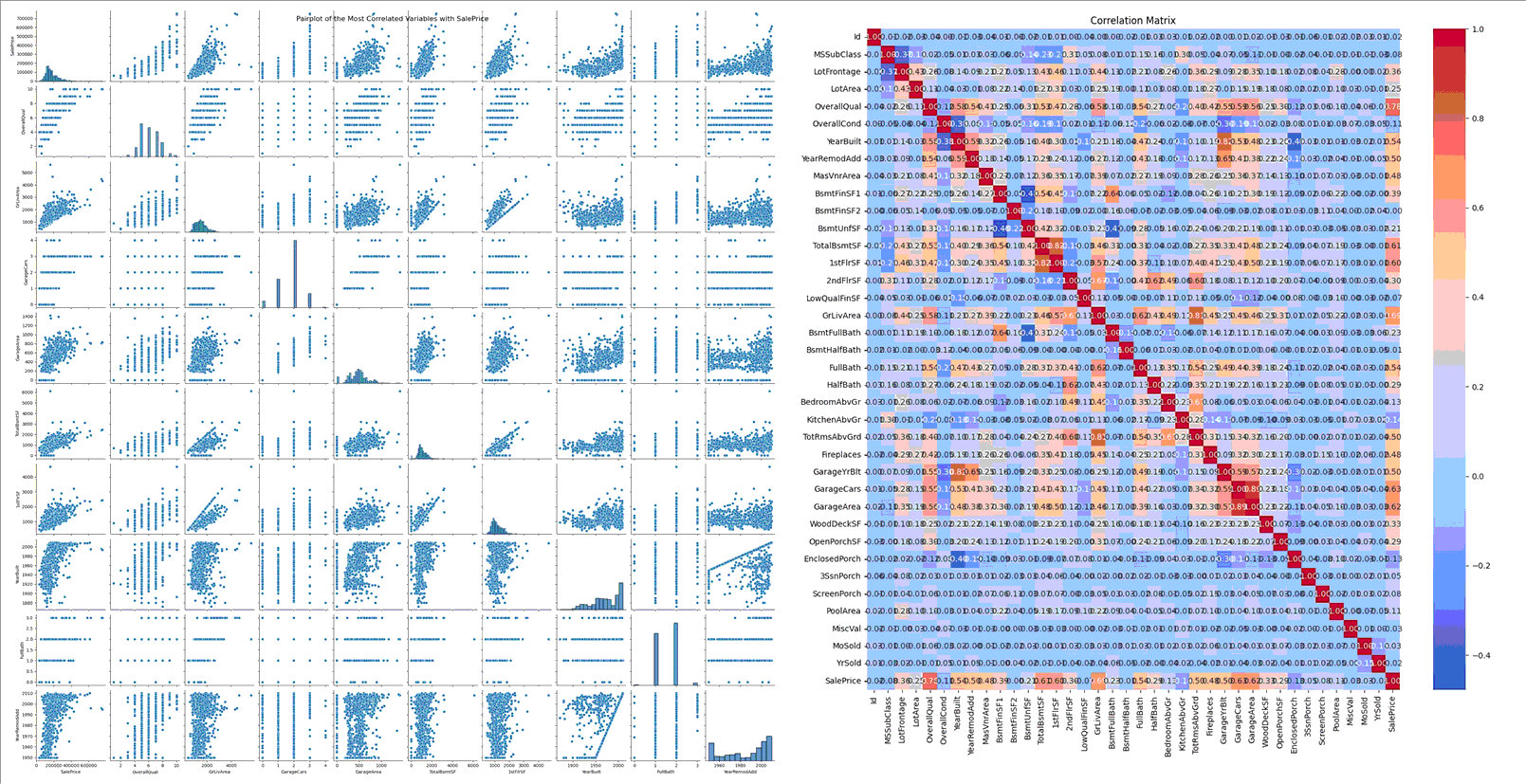}
    \caption{Data analysis and visualization capabilities of \model{}}
    \label{fig:visual}
\end{figure*}

\Cref{fig:visual} illustrates the results of data analysis and visualization of the \model{}.

\clearpage

\section{Details of datasets}
\subsection{Open-ended task details}
\label{sec:open_ended}
\begin{figure}[ht]
    \centering
        \includegraphics[width=0.9\textwidth]{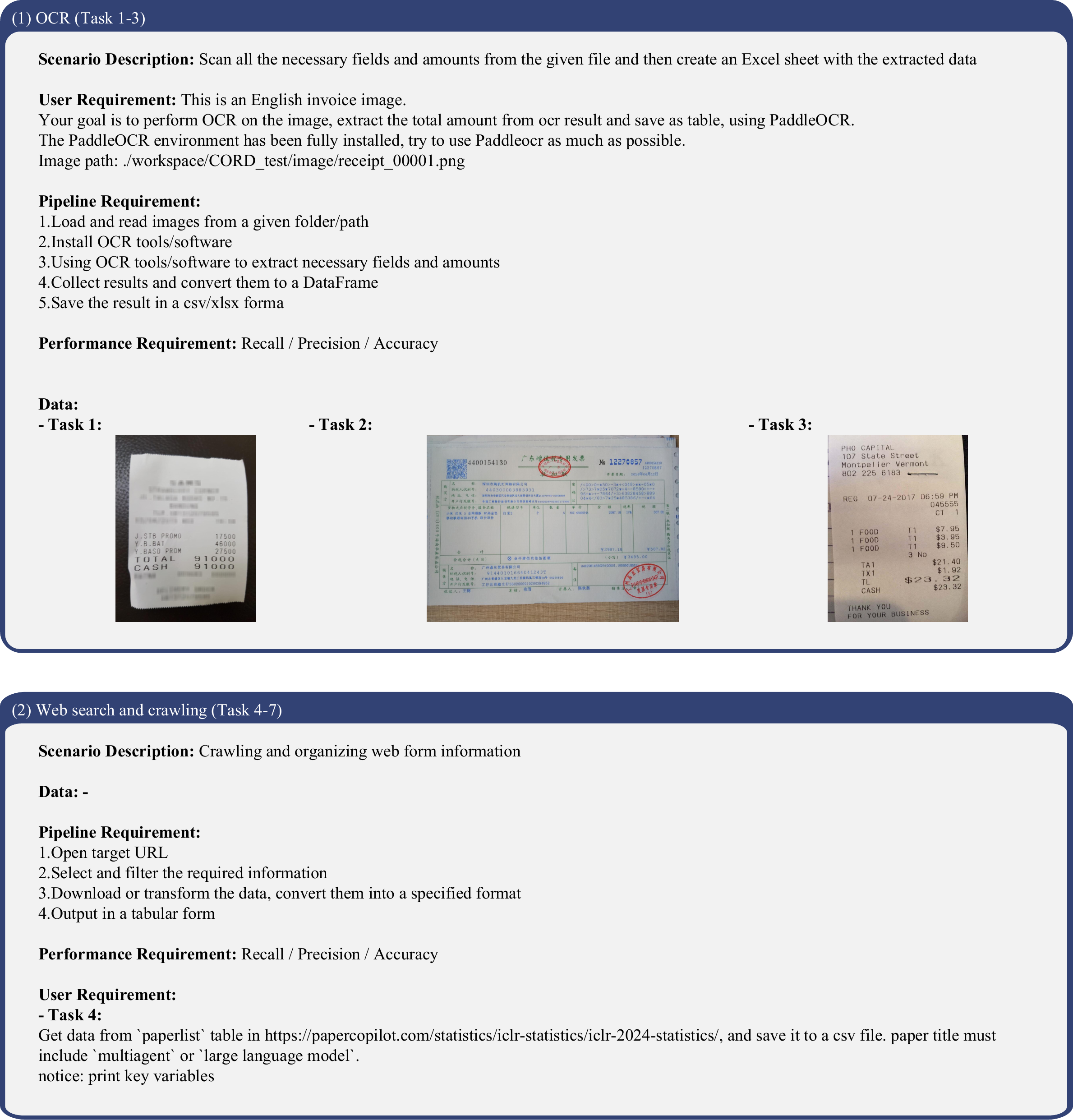}
            \caption{\textbf{Open-ended task cases (OCR and web search and crawling)} We present task 4, omitting similar tasks for brevity.}
    \label{fig:open_ended_task_details_task_1_7}
\end{figure}

\begin{figure}[t]
    \centering
        \includegraphics[width=0.9\textwidth]{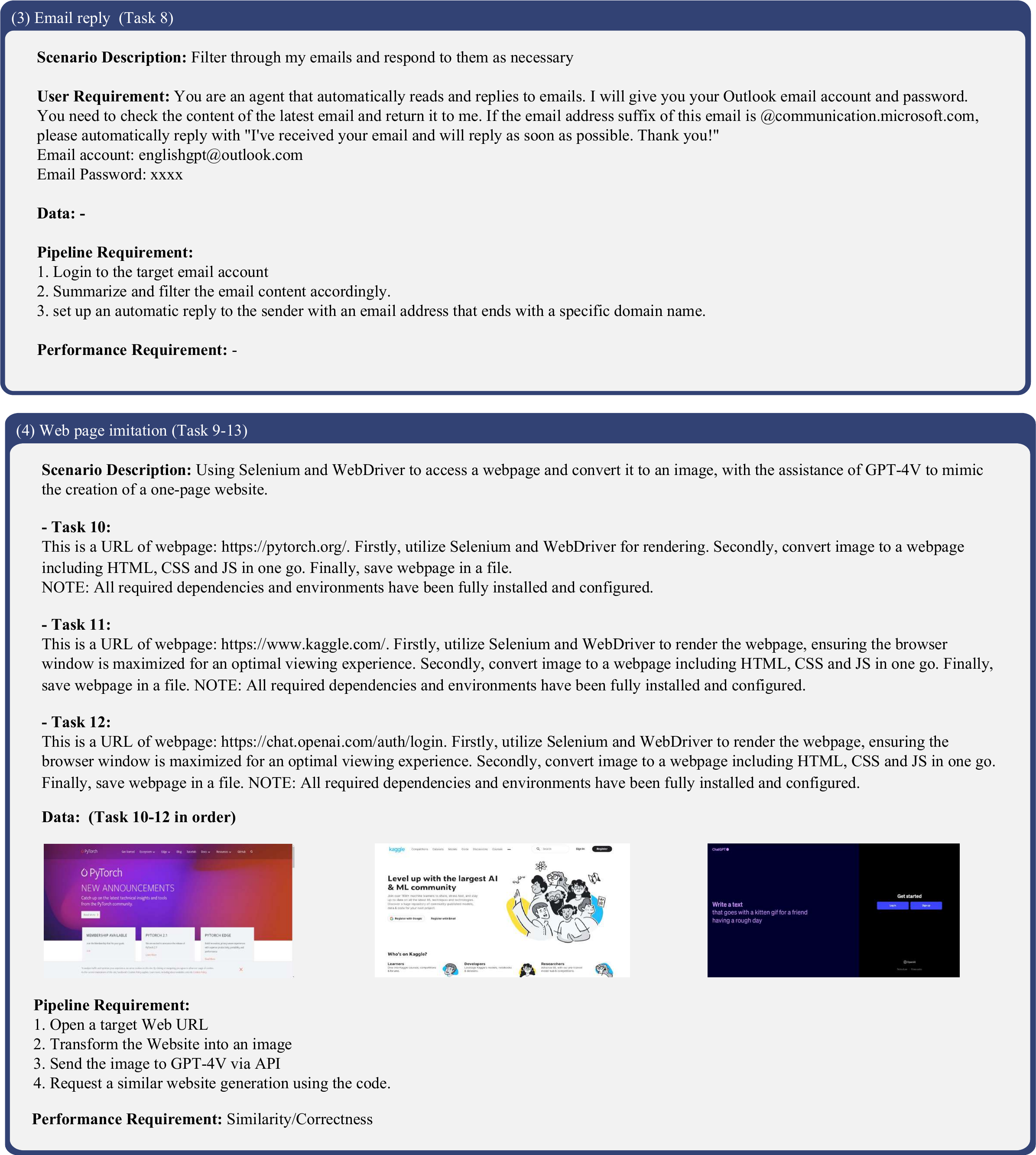}
            \caption{\textbf{Open-ended task cases (email reply and web page imitation).} We present tasks 10-12, omitting similar tasks for brevity.}
    \label{fig:open_ended_task_details_task_8_13}
\end{figure}

\begin{figure}[t]
    \centering
        \includegraphics[width=0.9\textwidth]{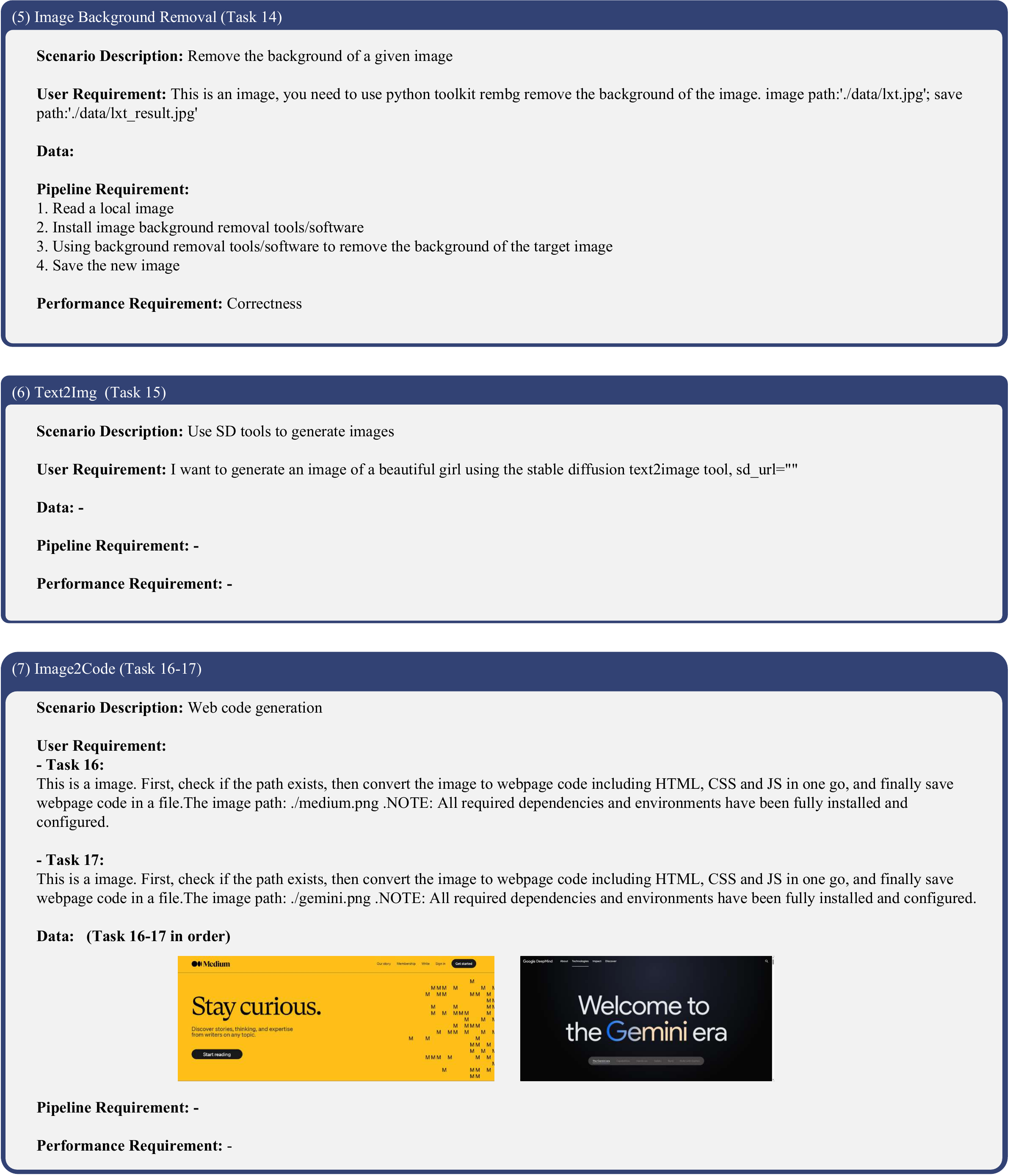}
            \caption{\textbf{Open-ended task cases (image background removal, text-to-image, and image-to-code)}}
    \label{fig:open_ended_task_details_task_14_17}
\end{figure}

\begin{figure}[t]
    \centering
        \includegraphics[width=0.9\textwidth]{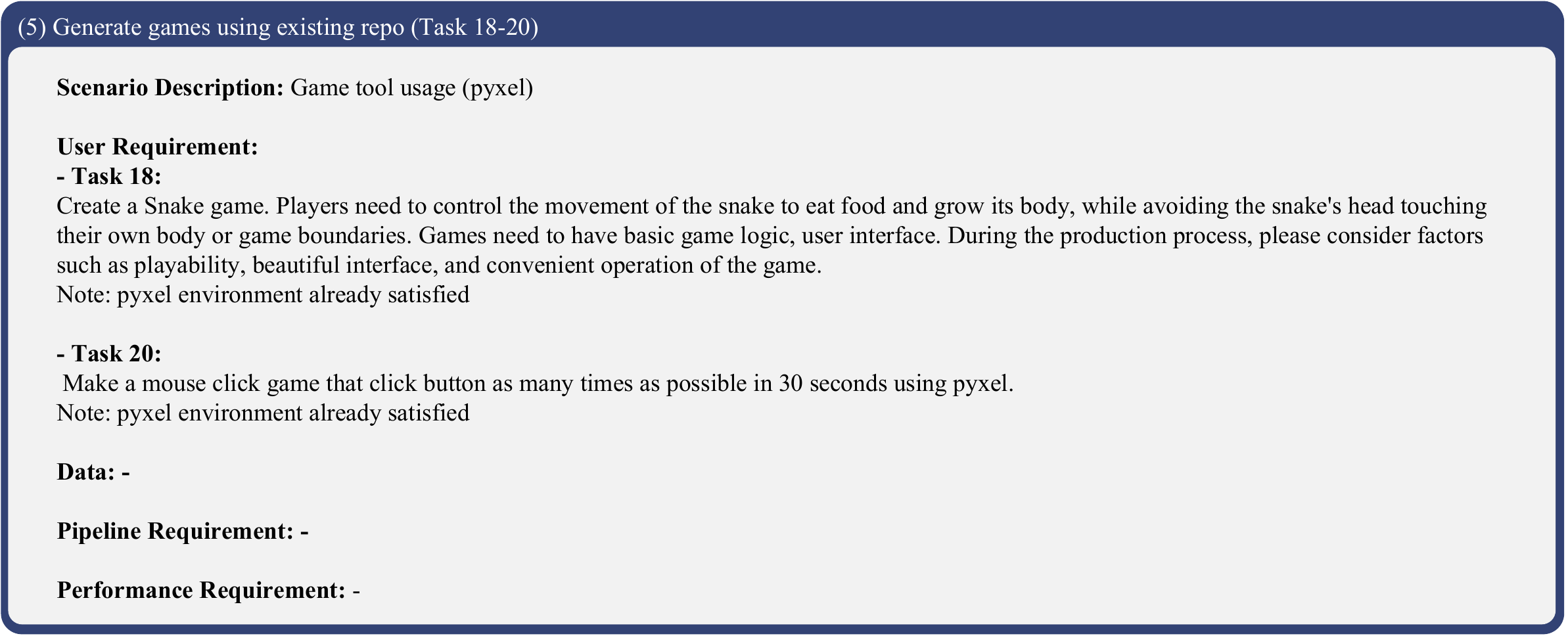}
            \caption{\textbf{Open-ended task cases (mini-game generation)} We present tasks 18 and 20, omitting similar tasks for brevity.}
    \label{fig:open_ended_task_details_task_18_20}
\end{figure}

\Cref{fig:open_ended_task_details_task_1_7,fig:open_ended_task_details_task_8_13,fig:open_ended_task_details_task_14_17,fig:open_ended_task_details_task_18_20} showcase several typical open-ended tasks in the following illustrations. For each task, we include the necessary data, user requirements, and assessment pipeline.

\subsection{ML-Benchmark dataset description}\label{appendix: ml-benchmark-dsp}
Here are the details about the ML-Benchmark dataset. We collect several typical datasets from Kaggle\footnote{https://www.kaggle.com/} and machine learning. Details are in \Cref{tab:dataset-info}

\begin{landscape}
\begin{table*}[t]
\centering
\caption{Details of the ML-Benchmark dataset, including dataset name, description, standard user requirements, dataset type, task type, difficulty, and metric used.}
\label{tab:dataset-info}
\footnotesize
\centering
\resizebox{0.94\linewidth}{!}{
\begin{tabular}{cc>{\arraybackslash}p{11cm}c>{\raggedright\arraybackslash}p{2.8cm}ccc} 
\toprule
ID & Dataset Name & User Req. & Dataset Type & Dataset Description & Task Type & Difficulty & Metric \\
\midrule
01 & Iris & Run data analysis on sklearn Iris dataset, including a plot & Toy & Suitable for EDA, simple classification and regression & EDA & 1 & \\
02 & Wine recognition & Run data analysis on sklearn Wine recognition dataset, include a plot, and train a model to predict wine class with 20\% as test set, and show prediction accuracy & Toy & Suitable for EDA, simple classification and regression & Classification & 1 & ACC \\
\midrule
03 & Breast Cancer & Run data analysis on sklearn Wisconsin Breast Cancer dataset, include a plot, train a model to predict targets (20\% as validation), and show validation accuracy & Toy & Suitable for EDA, binary classification to predict benign or malignant & Classification & 1 & ACC \\
\midrule
04 & Titanic & This is a Titanic passenger survival dataset, and your goal is to predict passenger survival outcomes. The target column is Survived. Perform data analysis, data preprocessing, feature engineering, and modeling to predict the target. Report accuracy on the eval data. Train data path: 'dataset\textbackslash{}titanic\textbackslash{}split\_train.csv', eval data path: 'dataset\textbackslash{}titanic\textbackslash{}split\_eval.csv'. & Beginner & Binary classification of survival, single table & Classification & 2 & ACC \\
\midrule
05 & House Prices & This is a house price dataset, and your goal is to predict the sale price of a property based on its features. The target column is SalePrice. Perform data analysis, data pre-processing, feature engineering, and modeling to predict the target. Report RMSE between the logarithm of the predicted value and the logarithm of the observed sales price on the eval data. Train data path: 'dataset\textbackslash{}house-prices-advanced-regression-techniques\textbackslash{}split\_train.csv', eval data path: 'dataset\textbackslash{}house-prices-advanced-regression-techniques\textbackslash{}split\_eval.csv'. & Beginner & Predicting house prices through property attributes, regression, single table & Regression & 2 & RMSLE \\
\midrule
06 & Santander Customer & This is a customer's financial dataset. Your goal is to predict which customers will make a specific transaction in the future. The target column is the target. Perform data analysis, data preprocessing, feature engineering, and modeling to predict the target. Report AUC on the eval data. Train data path: 'dataset\textbackslash{}santander-customer-transaction-prediction\textbackslash{}split\_train.csv', eval data path: 'dataset\textbackslash{}santander-customer-transaction-prediction\textbackslash{}split\_eval.csv' . & Industry & Binary classification to predict customer transactions, single table & Classification & 2 & AUC \\
\midrule
07 & ICR - Identifying & This is a medical dataset with over fifty anonymized health characteristics linked to three age-related conditions. Your goal is to predict whether a subject has or has not been diagnosed with one of these conditions. The target column is Class. Perform data analysis, data preprocessing, feature engineering, and modeling to predict the target. Report F1 Score on the eval data. Train data path: 'dataset\textbackslash{}icr-identify-age-related-conditions\textbackslash{}split\_train.csv', eval data path: 'dataset\textbackslash{}icr-identify-age-related-conditions\textbackslash{}split\_eval.csv' . & Industry & Binary classification of health symptoms, single table & Classification & 2 & F1 \\
\midrule

08 & Santander Value & This is a customer's financial dataset. Your goal is to predict the value of transactions for each potential customer. The target column is the target. Perform data analysis, data preprocessing, feature engineering, and modeling to predict the target. Report RMSLE on the eval data. Train data path: 'dataset\textbackslash{}santander-value-prediction-challenge\textbackslash{}split\_train.csv', eval data path: 'dataset\textbackslash{}santander-value-prediction-challenge\textbackslash{}split\_eval.csv' . & Industry & Predicting transaction values, regression, single table, 5k columns, suitable for complex algorithms & Regression & 3 & RMSLE \\
\bottomrule
\end{tabular}
}
\end{table*}
\end{landscape}

\end{document}